\documentclass[final,12pt]{elsarticle}
\input{preamble}
\usepackage[usenames,dvipsnames]{xcolor}
\usepackage[para,online,flushleft]{threeparttable}
\biboptions{longnamesfirst,angle,semicolon}
\usepackage[utf8]{inputenc}
\usepackage{tikz}
\usetikzlibrary{shadows.blur}
\usepackage{algorithm}
\usepackage[noend]{algorithmic}
\usepackage[T1]{fontenc}
\usepackage{comment}
\usepackage{xspace}
\usepackage{xfrac}
\usepackage{mathtools}
\usepackage{siunitx}
\usepackage{wrapfig}

\usepackage[belowskip=-3pt,aboveskip=1.5pt]{caption}
\usepackage{subcaption}
\usepackage{float}
\usepackage{pifont}
\usepackage{adjustbox}
\usepackage[capitalize]{cleveref}

\definecolor{Gold}{rgb}{1,0.84,0}
\definecolor{Bronze}{rgb}{0.69,0.55,0.34}
\newcommand{\cmark}{{\color{OliveGreen}\ding{51}}}

\newcommand{\xmark}{{\color{BrickRed}\ding{55}}}
\usetikzlibrary{calc, positioning, arrows, decorations.pathreplacing, decorations.pathreplacing}
\setcitestyle{square}

\let\Algorithm\algorithm
\renewcommand\algorithm[1][]{\Algorithm[#1]\setstretch{1.2}}
\begin{document}
\setlength{\abovedisplayskip}{3pt}
\setlength{\belowdisplayskip}{3pt}
\linepenalty=1000
 \glsdisablehyper
\begin{frontmatter}

\title{\bf
\gosafeopt: Scalable Safe Exploration \\for Global Optimization of Dynamical Systems
}
\author[1]{\bf Bhavya Sukhija}
\ead{bhavya.sukhija@inf.ethz.ch}
\author[1]{\bf Matteo Turchetta}
\ead{matteo.turchetta@inf.ethz.ch}
\author[1]{\bf David Lindner}
\ead{david.lindner@inf.ethz.ch}
\author[1]{\bf Andreas Krause}
\ead{krausea@ethz.ch}
\author[3]{\bf Sebastian Trimpe}
\ead{trimpe@dsme.rwth-aachen.de}
\author[4,5]{\bf Dominik Baumann}
\ead{dominik.baumann@aalto.fi}
\address[1]{Department of Computer Science, ETH Z\"urich, Switzerland}
\address[3]{Institute for Data Science in Mechanical Engineering, RWTH Aachen University, Germany}
\address[4]{Department of Electrical Engineering and Automation, Aalto University, Espoo, Finland}
\address[5]{Department of Information Technology, Uppsala University, Sweden}

\begin{abstract}
\looseness=-1
    Learning optimal control policies directly on physical systems is challenging. 
    Even a single failure can lead to costly hardware damage. Most existing model-free learning methods that guarantee safety, i.e., no failures, during exploration are limited to local optima. 
    This work proposes \gosafeopt as the first provably safe and optimal algorithm that can safely discover globally optimal policies for systems with high-dimensional state space. 
    We demonstrate the superiority of \gosafeopt over competing model-free safe learning methods in simulation and hardware experiments on a robot arm. 
\end{abstract}

\begin{keyword}
Model-free learning, Bayesian Optimization, Safe learning.
\end{keyword}

\end{frontmatter}

\section{Introduction}\label{intro}
\looseness=-1
The increasing complexity of modern dynamical systems often makes deriving mathematical models for
traditional model-based control approaches forbiddingly involved and time-consuming. Model-free \gls{RL} methods~\cite{Sutton1998} are a promising alternative 
as they learn control policies directly from data. To succeed, they need to explore the system and its environment. Without a model, this can be risky and unsafe. Since modern hardware such as robots are expensive and their repairs are time-consuming, safe exploration is crucial to apply model-free \gls{RL} in real-world problems.
This paper proposes \gosafeopt, a model-free learning algorithm that can search for globally optimal policies while guaranteeing safe exploration with high probability. 
\subsection{Related Work}\label{sec:related_work}
\looseness -1
Advances in machine learning have motivated the usage of model-free \gls{RL} algorithms for obtaining control policies
~\cite{levine2016end,peters2008reinforcement,lillicrap2015continuous,kober2013reinforcement,schaal2010learning}.
However, directly applying these methods to  policy optimization presents two major challenges: \emph{(i)} Machine learning algorithms often require large amounts of data. 
In learning control, such data is often gathered by conducting experiments with physical systems, which is time-consuming and wears out the hardware.
\emph{(ii)} Learning requires exploration, which can lead to unwarranted and unsafe behaviors. 

Challenges \emph{(i)} and \emph{(ii)} can be addressed jointly by \gls{BO} with constraints. \gls{BO}~\cite{mockus1978application} is a class of black-box global optimization algorithms, that has been used in a 
variety of works~\cite{Calandra2016,DBLP:journals/corr/MarcoHBST16,DBLP:journals/corr/AntonovaRA17,DBLP:journals/corr/abs-1910-13399} to optimize controllers in a sample-efficient manner.
In constrained \gls{BO}, there are two main classes of methods. 
On the one hand, approaches like~\cite{Gelbart,hernandez2016general,classRegress, heim2020learnable} find safe solutions but allow unsafe evaluations during training.
Herein, we focus on approaches that guarantee safety at all times during exploration, which is crucial when dealing with expensive hardware.
\safeopt~\cite{pmlr-v37-sui15} and safe learning methods that emerged from it, e.g.,~\cite{DBLP:journals/corr/BerkenkampSK15,DBLP:journals/corr/BerkenkampKS16, koenig2021safe}, guarantee safe exploration with high probability by exploiting properties of the constraint functions, e.g., regularity.
Unfortunately, these methods are limited to exploring a safe set connected with a known initial safe policy. Therefore, they could miss the global optimum in the presence of disjoint safe regions in the policy space (see ~\cref{fig:disjoint_sets_illustration}). Disjoint safe regions appear when learning an impedance controller for a robot arm, as we show in our experiments and in many other applications~\cite{GRYAZINA200613,Calandra2016,DBLP:gosafe}.
To address this limitation \cite{DBLP:gosafe} proposes \gosafe, which can provably and safely discover the safe global optimum in the presence of disjoint safe regions under mild conditions. To achieve this, it learns safe backup policies for different states and uses them to preserve safety when evaluating policies outside of the safe set.  Specifically, it switches between actively exploring local safe regions in the state and policy space and safe global exploration. 
However, the active exploration in the state and policy space requires a coarse discretization of the space and is
infeasible for all but the simplest systems with  low-dimensional state spaces, \cite{DBLP:journals/corr/abs-1902-03229} argues that dimension $d> 3$ is already challenging. As a result, \gosafe cannot only handle most real-world dynamical systems, and is restricted to impractical systems with low-dimensional state spaces. 
The concept of switching between two exploration stages is also pursued in
the stagewise safe optimization algorithm proposed in~\cite{sui2018stagewise}.
However, also~\cite{sui2018stagewise} is restricted to an optimum connected to a safe initialization. 
Lastly, the general idea of learning backup policies is related to safety filters and control barrier functions~\cite{wabersich2021predictive,wieland2007constructive,cheng2019end}.
Nevertheless, those methods require either availability or learning of a dynamics model besides learning the policy and are, therefore, model-based.
In this work, we focus on a model-free approach.

\subsection{Contributions}
\begin{figure}[t]
\centering
\begin{minipage}{0.49\textwidth}
   \centering
   \includegraphics[width=\textwidth]{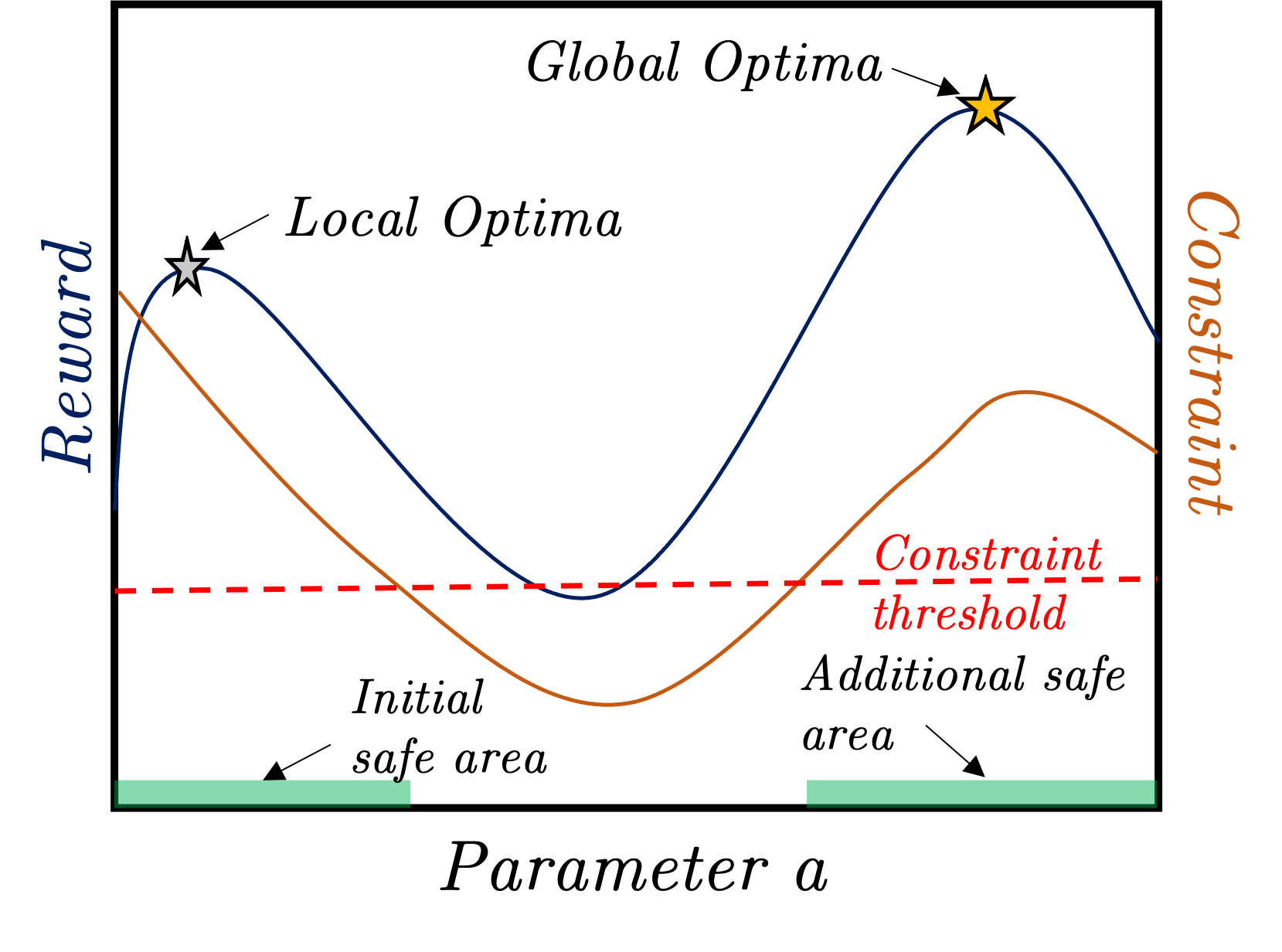}
   \caption{Illustrative example with disjoint safe regions in the policy space. \capt{The blue line depicts the objective, and the orange line is the constraint function. 
   There are two safe regions that are marked in green. \safeopt cannot explore the global optimum if it is initialized in the left region. }}
    \label{fig:disjoint_sets_illustration}
\end{minipage}
\begin{minipage}{0.49\textwidth}
   \vspace{-6.5em}
   \centering
   \includegraphics[width=0.7\textwidth]{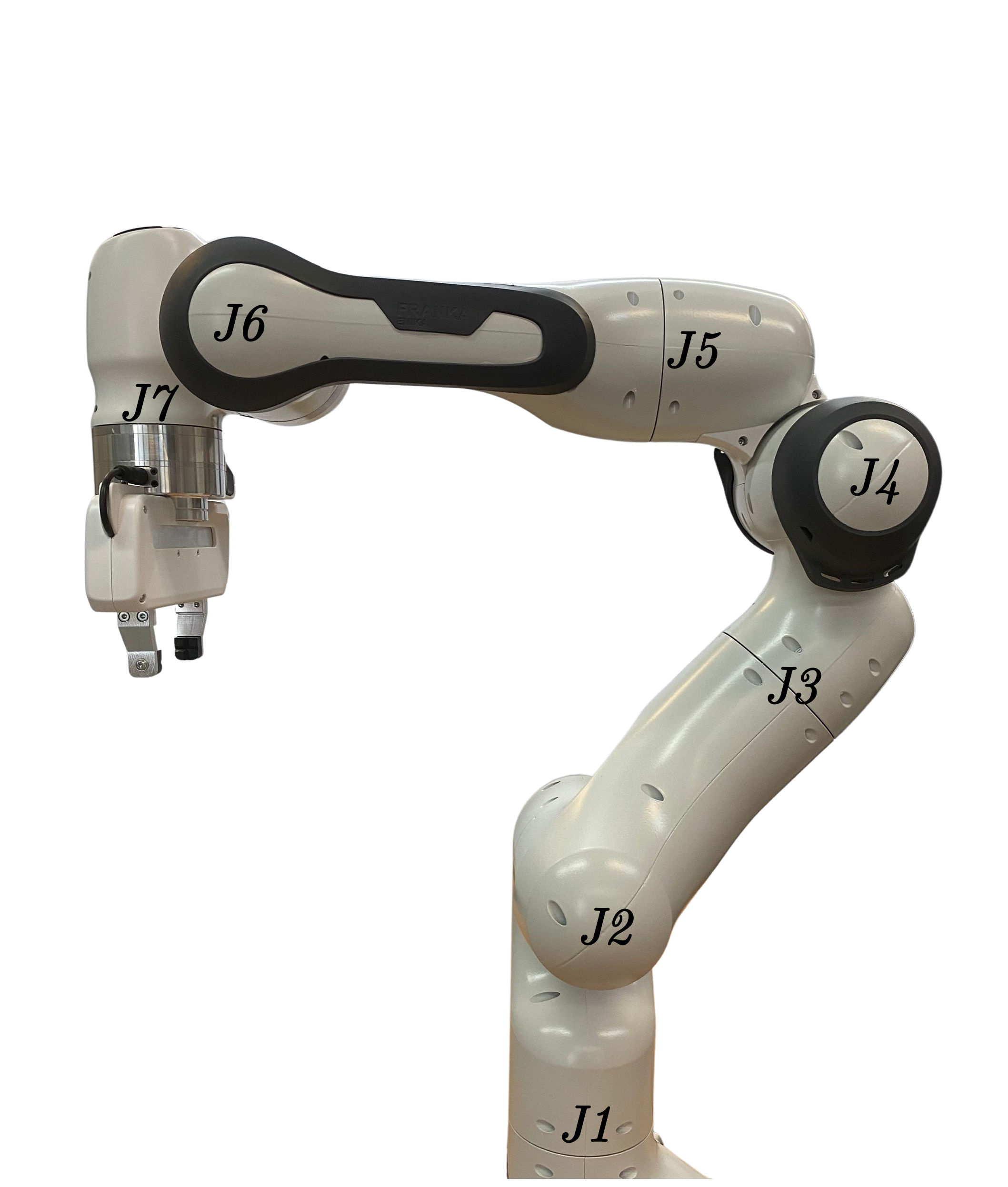}
   \captionsetup{width=.9\linewidth}
   \caption{Franka Emika Panda; seven degrees of freedom robot arm used for our evaluations. }
    \label{fig:franka_panda}
\end{minipage}
\end{figure}

This work presents \gosafeopt, the first model-free algorithm that can globally search optimal policies for safety-critical, real-world dynamical systems, i.e., systems with high-dimensional state spaces. 
\gosafeopt does not discretize and actively explores the state space. Therefore, it overcomes the main shortcomings and restrictions of \gosafe, while still performing safe global exploration. This makes \gosafeopt the first and only model-free safe global exploration algorithm for real-world dynamical systems.
Crucially, \gosafeopt leverages  the \textit{Markov property} of the system's state to learn backup policies which it uses to guarantee safety when evaluating policies outside the safe set. This novel mechanism for learning backup policies does not depend on the dimension of the state space. 
We provide high-probability safety guarantees for \gosafeopt and we prove that it recovers the safe globally optimal policy under assumptions that hold for many practical cases. 
Finally, we validate it in both simulated and real safety-critical path following experiments on a robotic arm (see \cref{fig:franka_panda}), which is prohibitive for \gosafe, the only competing  model-free global safe search method. Further, we show that \gosafeopt achieves considerably better performance than \safeopt, a state-of-the-art method for local model-free safe policy search, and its high-dimensional variants.
\cref{tab:prior_work} compares \gosafeopt to \safeopt and \gosafe in terms of safety guarantees, scalability, global exploration, and sample efficiency. It shows that \gosafeopt is the only method that can perform sample-efficient global exploration in high-dimensional systems while providing safety guarantees.
\begin{table}[t]
\newcommand{\extension}{\xmark\tnote{1}}
\centering
\caption{\small Comparison of \gosafeopt and prior work on safe exploration based on their safety guarantees, scalability, global exploration, and sample efficiency. } 
\begin{adjustbox}{max width=\linewidth}\begin{threeparttable}
    \begin{tabular}{c|cccc}
       \toprule
       \multirow{2}{*}
        & Safe exploration &
        State space with   &
        Global exploration &  
        Sample efficient \\
        & & dimension $d>3$ & & \\
        \midrule
        \textbf{\safeopt}~\cite{DBLP:journals/corr/BerkenkampKS16} & \cmark & \cmark & \xmark & \cmark \\
        \textbf{\gosafe}~\cite{DBLP:gosafe} & \cmark & \xmark & \cmark & \xmark \\
        \textbf{\gosafeopt (ours)} & \cmark & \cmark & \cmark & \cmark \\
       \bottomrule
    \end{tabular}
\end{threeparttable}\end{adjustbox}
\label{tab:prior_work}
\end{table}

\section{Problem Setting}\label{sec:problem_setting}
We consider a Lipschitz-continuous system 
\begin{equation}
    \diff x(t) = z(x(t),u(t))\diff t,
    \label{eq:syseq}
\end{equation}
where $z(\cdot)$ represents the unknown system dynamics, \gls{statevec} $\in \gls{statespace} \subset \R^{s}$ is the system state and \gls{input} $\in \mathcal{U} \subset \R^{p}$ is the input we apply to steer the system state to follow a desired trajectory \gls{statedes} $\in \gls{statespace}$ for all $t \geq 0$. We assume that the system starts at a known initial state $x(0)=\gls{x0}$. 

The control input $u(t)$ we apply for a given state $x(t)$ is specified by a policy $\pi: \mathcal{X}\times\mathcal{A}\to \mathcal{U}$, with $u(t)=\pi\left(x(t),a\right)\coloneqq\pi^a(x(t))$. 
The policy is parameterized by $ \gls{policyparam} \in \mathcal{A} \subset \R^d$, where $\mathcal{A}$ is a finite parameter space\footnote{infinite parameter spaces can be handled via discretization (e.g., random subsampling)}. 
We encode our goal of following the desired trajectory \gls{statedes} through an objective function, $f: \gls{actionspace} \to \R$. 
Note, the trajectory of a deterministic system~\eqref{eq:syseq} is fully determined by its initial state $x_0$ and the control policy. Therefore, the objective is independent of the state space $\mathcal{X}$.
We seek for a controller parametrization $a\in\mathcal{A}$ that optimizes $f$ for a constant initial condition $x_0$.
Since the dynamics of the system in \cref{eq:syseq} is unknown, so is the objective $f$. 
Nonetheless, we assume we obtain a noisy measurement of $f(a)$ at any $a\in\mathcal{A}$ by running an experiment.
We aim at optimizing $f$ from these measurements in a sample-efficient way. 
Additionally, to avoid the deployment of harmful policies,
we formulate safety as a set of \emph{unknown} constraints over the system trajectories that must be satisfied \emph{at all times}. 
Similar, as for $f$, these constraints only depend on the parameter $a$ and hence take the form $g_i : \gls{actionspace} \to \R$ for each constraint function $g_i$, where $i\in\{1,\ldots,q\}\coloneqq \mathcal{I}_g$ and $q\in\mathbb{N}$. 
The resulting constrained optimization problem with unknown objective and constraints is:
\begin{equation}
    \max_{a \in \gls{actionspace}} f(a) \quad \text{subject to} \quad g_i(a) \geq 0, \forall i \in \mathcal{I}_g.
    \label{eq:problem_statement}
\end{equation}
We represent the objective and constraints using a scalar-valued function in a higher dimensional domain, proposed by~\cite{DBLP:journals/corr/BerkenkampKS16}: 
\begin{equation}
    h(a,i)=\begin{cases}
    f(a) & \text{if } i=0, \\
    g_i(a) & \text{if } i \in \gls{conset},
    \end{cases}
    \label{h_function}
\end{equation}
with $\gls{conset}=\conset$, $\mathcal{I}\coloneqq\{0,1,\dots,q\}$, and $i\in \mathcal{I}$. This representation will later help us in learning the unknown function. 

In summary, \emph{our goal is to find the optimal and safe policy parameter for the system starting from the nominal initial condition $x_0$}.
We refer to the solution of \cref{eq:problem_statement} as the safe global optimum $a^*$.  Note, finding the optimal policy for a fixed initial condition $x_0$ is a common task in episodic \gls{RL}~\cite{Sutton1998}.

Solving this problem without a dynamics model and without incurring failures for generic systems, objectives, and constraints is hopeless. The following section introduces our assumptions to make this problem tractable.

\subsection{Assumptions}\label{sec:accumption}
To solve the problem in \cref{eq:problem_statement} safely, we assume to have at least one initial safe policy to start data collection without violating constraints. 
This initial policy could be derived from available simulators, first principles models, or by performing controlled experiments on the hardware directly. This policy can be conservative and sub-optimal. For instance, for mobile robots, a policy that barely moves the robot could be an initial safe policy.
\begin{assumption}
A set $S_0\subset\mathcal{A}$ of safe parameters is known. That is, for all parameters $a$ in $S_0$ we have $g_i(a)\geq 0$ for all $i\in\mathcal{I}_g$.
\label{ass:safe_seed}
\end{assumption}

In practice, similar policies often lead to similar outcomes. In other words, the objective and the constraints exhibit regularity properties.
We capture this by assuming that the function $h$,~\cref{h_function}, lives in an \gls{RKHS}~\cite{SchoelkopfKernels} and has bounded norm in that space.
\begin{assumption}
    The function $h$ lies in an \gls{RKHS} associated to a kernel $k$ and has a bounded norm in that RKHS $\|h\|_k\leq B$. Furthermore, the objective $f$ and constraints $g_i$ are Lipschitz continuous with known constants.
    \label{ass:smoothness_assumption}
\end{assumption}
Without Assumption~\ref{ass:smoothness_assumption}, the constraint and reward functions can be discontinuous making it impossible to infer the safety of a policy before evaluating it and to provide safety guarantees. In practical applications, such behavior is undesirable, and therefore rare. For further discussion on the practicality of this assumption, we refer the reader to~\cite{tight_bounds}.

Next, we formalize our assumptions on the measurement model.
\begin{assumption}
We obtain noisy measurements of $h$ with the  
measurement noise \gls{iid} $\sigma$-sub-Gaussian. 
That is, for a measurement $y_i$ of $h(\cdot, i)$, we have $y_i = h(a, i) + \epsilon_i$ with $\epsilon_i$ $\sigma$-sub-Gaussian for all $i\in \mathcal{I}$.
\label{ass:observation_model}
\end{assumption}

Assumptions \ref{ass:safe_seed}, \ref{ass:smoothness_assumption}, and \ref{ass:observation_model} are common in the safe \gls{BO} literature~\cite{DBLP:journals/corr/BerkenkampSK15,DBLP:journals/corr/BerkenkampKS16,pmlr-v37-sui15}. However, these approaches treat the evaluation of a policy as a black box. In contrast, we monitor the rollout of a policy to intervene and bring the system back to safety, if necessary. This can be achieved for a Markovian~\cite{puterman2014markov} system, like the one we consider in~\cref{eq:syseq} (see~\cref{prop:backup_policies} in the appendix).

To monitor the rollouts, we assume that we receive a state measurement after every $\Delta t$ seconds and that in between discrete time steps, the system cannot arbitrarily jump, i.e., its movement within these (typically small) time intervals is bounded. Note, for many robotic systems this assumption is valid. Especially, since we can choose the sampling time $\Delta t$. However, estimating this bound can be challenging. A conservative value for the bound may be estimated by performing controlled experiments, e.g, with the safe initial policy from Assumption~\ref{ass:safe_seed}, directly on hardware. Simulators or first principle models, if available, can also be leveraged.
\begin{assumption}
The state $x(t)$ is measured after every $\Delta t$ seconds. Furthermore, 
for any $x\idxk$ and $\rho \in [0,1]$, the distance to $x(t+\rho\Delta t)$ induced by any action is bounded by a known constant $\Xi$, that is, 
$\normAny{x(t+\rho\Delta t) - x\idxk} \leq \Xi$. 
\label{ass:one_step_jump}
\end{assumption}
\extranote Implicitly, we here assume noise-free measurements of the state for simplicity. Our method also works for the noisy case (see~\cref{noisy_case_bc}), which is typical in the real world.

Triggering a backup policy for a Markovian system is not sufficient to guarantee the safety of the whole trajectory for a generic constraint. Consider the case where safety is expressed as a constraint on a cost accumulated along the trajectory. Even if we are individually safe before and after triggering a backup policy, we might be unsafe overall. 
Therefore, we limit the types of constraints we consider.
\begin{assumption}
We assume that, for all $i\in\{1,\ldots,q\}$, $g_i$ is defined as the minimum of a state-dependent function $\bar{g}_i$ along the trajectory starting in $x_0$ with controller $\pi^a$.
Formally:
\begin{equation}
    g_i(a) = \min_{x' \in \xi_{(0,x_0,a})} \bar{g}_i(x'),
\end{equation}
with $\xi_{(0,x_0,a)}\coloneqq \{x_0+\int_0^tz(x(\tau);\pi^a(x(\tau))\mathrm{d}\tau\}$ the trajectory of $x(t)$ under policy parameter $a$ starting from $x_0$ at time $0$.
\label{ass:constraint}   
\end{assumption}
An example of such a constraint is the minimum distance of the system to an obstacle.
We can now provide a formal definition of a safe experiment.
\begin{definition}
An experiment is safe if, for all $t \geq 0$ and all $i\in\{1, \dots, q\}$,
\begin{equation}
    \bar{g}_i(x(t)) \geq 0.
\end{equation}
\label{def:safe_experiment}
\end{definition}
\vspace{-2em}
This is a more general way of defining safety for the optimization problem from~\cref{eq:problem_statement}. In particular, where~\cref{eq:problem_statement} only considers trajectories associated with a fixed policy parameter $a$, \cref{def:safe_experiment} also covers the case in which different portions of the trajectory are induced by different controllers. 
\section{Preliminaries}\label{sec:preliminaries}
\glsreset{GP}
\looseness=-1
This section reviews \glspl{GP} and how to use them to construct frequentist confidence intervals, as well as relevant prior work on safe exploration (\safeopt).
\subsection{Gaussian Processes} \label{sec:GP}
We model our unknown objective and constraint functions using \gls{GPR}~\cite{Rassmussen}. 
In \gls{GPR}, our prior belief is captured by a \gls{GP}, which is fully determined by a prior mean function\footnote{Assumed to be zero \gls{wlog}.} and a covariance function $k\left(a,a'\right)$.
Importantly, if the observations are corrupted  by \gls{iid} Gaussian noise with variance $\sigma^2$, i.e., $y_i = f(a_i) + v_i$, and $v_i \sim \N(0,\sigma^2)$, the posterior over $f$ is also a \gls{GP} whose mean and variance can be computed in closed form.  Let us denote with $Y_n \in \R^n$ the array containing $n$ noisy observations of $f$, then the posterior  of $f$ at $\Bar{a}$ is $f\left(\Bar{a}\right) \sim \mathcal{N}\left(\mu_{n} \left(\Bar{a}\right),\sigma^2_{n}\left(\Bar{a}\right)\right)$ where
\begin{subequations}
\label{eq:kernel_posterior}
\begin{align}
    \mu_n\left(\Bar{a}\right) &= k_n\left(\Bar{a}\right)(K_n+I_n\sigma^2)^{-1}Y_n, \\
    \sigma^2_n\left(\Bar{a}\right)  &= k\left(\Bar{a},\Bar{a}\right)-k_n\left(\Bar{a}\right)(K_n+I_n\sigma^2)^{-1}k_n^T \left(\Bar{a}\right).
\end{align}
\end{subequations}
The entry $(i,j) \in \{1,\dots,n\} \times \{1,\dots,n\}$ of the covariance matrix $K_n\in\R^{n\times n}$ is $k\left(a_i,a_j\right)$, $k_n\left(\Bar{a}\right)=[k(\Bar{a},a_1),\ldots,k(\Bar{a},a_n)]$ captures the covariance between  $x^*$ and the data, and $I_n$ is the $n\times n$ identity matrix.

\cref{eq:kernel_posterior} considers the case where $f$ is a scalar function. 
To model the objective $f$ \emph{and} constraints $g_i$, we use the selector function from~\cref{h_function}.

\subsection{Frequentist Confidence Intervals} \label{conf_intervals}
To avoid failures, we must determine the safety of a given policy before evaluating it. To this end, we reason about plausible worst-case values of the constraint $g_i$ for a new policy $a$. We use the posterior distribution over the objective and constraints given by \cref{eq:kernel_posterior} to build frequentist confidence intervals that hold with high probability, i.e., at least $1-\delta$, and are of the form:
\begin{equation}
    |\mu_{n-1}(a, i)-h(a, i)| \leq \beta^{\sfrac{1}{2}}_n \sigma_{n-1}(a, i), \quad \forall i \in \mathcal{I}.
    \label{uncertainty_bound}
\end{equation}
\looseness=-1
For functions fulfilling Assumption \ref{ass:smoothness_assumption} and \ref{ass:observation_model}, \cite{srinivas,pmlr-v70-chowdhury17a} derive an appropriate value for $\beta_{n}$. This value depends on $\delta$, $n$ and
the maximum information gain $\gamma_{n}$, cf.,~\cite{elementsofIT}\footnote{The maximum information gain  is $\gamma_{n} \allowbreak \coloneqq \underset{A \subset D: |A|=n}{\max}I(y_A;f_A)$, where $I(y_A;f_A)$ is the mutual information between $f_A$ evaluated at points in $A$ and the observations $y_A$, that is the amount of information $y_A$ contains about $f$~\cite{elementsofIT}.}. 
\subsection{\safeopt for Model-Free Safe Exploration} \label{safeopt_gosafe}
\looseness=-1
\safeopt leverages the confidence intervals presented in \cref{conf_intervals} to solve black-box constrained optimization problems while guaranteeing safety for all the iterates with high probability.
It ensures safety by limiting its evaluations to a set of provably safe inputs. 
In particular, \safeopt defines the lower bound of the confidence interval $l_n$ as $l_n(a,i) = \max \{\allowbreak l_{n-1}(a,i), \allowbreak  \mu_{n-1}(a,i)-\beta^{\sfrac{1}{2}}_n\sigma_{n-1}(a,i)\}$, with $l_0(a,i) = 0$ for all $a\in S_0$, $i \in \gls{conset}$ and $-\infty$ otherwise, and the upper bound $u_n$ as $u_n(a,i) = \min \{\allowbreak  u_{n-1}(a,i), \allowbreak \mu_{n-1}(a,i)\allowbreak +\beta^{\sfrac{1}{2}}_n\sigma_{n-1}(a,i)\}$ with $u_0(a, i) = \infty$ for all $a\in \mathcal{A}, i \in \mathcal{I}$. 
Given a set of safe parameters $S_{n-1}$, it then infers the safety of nearby parameters by combining the confidence intervals with the Lipschitz continuity of the constraints:
\begin{equation}
    \label{eq:safeset}
     S_n \coloneqq \bigcap\limits_{i\in\mathcal{I}_g}\bigcup\limits_{a' \in S_{n-1}} \{a \in\mathcal{A} \mid l_n(a',i)-L_\mathrm{a}\normAny{a-a'} \ge 0\},
\end{equation}
with $L_a$ the joint Lipschitz constant of $f(a)$, $g_i(a)$.  This leads to a local expansion of the safe set.
Thus, in the case of disconnected safe regions, the optimum discovered by \safeopt may be local (see \cref{fig:disjoint_sets_illustration}).
\section{\gosafeopt}\label{sec:gosafeopt}
In this section, we present our algorithm, \gosafeopt, which combines the sample efficient local exploration of \safeopt with global exploration to safely discover globally optimal policies for dynamical systems. To the best of our knowledge, \gosafeopt is the first model-free algorithm that can globally search for optimal policies, guarantee safety during exploration, and is applicable to complex hardware systems.

\subsection{The algorithm}\label{sec:main_algo}
\gosafeopt consists of two alternating stages, \gls{LSE} and \gls{GE}. In \gls{LSE}, we 
explore the safe portion of the parameter space connected to our current estimate of the safe set.
Crucially, we exploit the Markov property to learn backup policies for each state we visit during \gls{LSE} experiments. 
During \textbf{GE}, we evaluate potentially unsafe policies in the hope of identifying new, disconnected safe regions. The safety of this step is guaranteed by triggering the backup policies learned during \gls{LSE} whenever necessary. If a new disconnected safe region is identified, we switch to a \textbf{LSE} step. Otherwise, \gosafeopt terminates and recommends the optimum $a^* = \underset{a \in S_n}{\arg\max}$ $l_n(a,0)$. 


In the following, we explain the \textbf{LSE} and \textbf{GE} stages in more detail and provide their pseudocode  in \cref{alg:LSE,alg:GE}, respectively. \Cref{alg:GSafeopt} presents the pseudocode for the full \gosafeopt algorithm. 

\subsubsection{Local Safe Exploration}
\looseness=-1
Similar to \safeopt, during \gls{LSE} we restrict our evaluations to provably safe policies, i.e., policies in the safe set, which is initialized with the safe seed from Assumption \ref{ass:safe_seed} and is updated recursively according to \cref{eq:safeset}
(line~\ref{lst:line:update_S_lse} in \cref{alg:LSE}).
We focus our evaluations on two relevant subsets of the safe set introduced in \cite{pmlr-v37-sui15}: the maximizers $\M$, i.e., plausibly optimal parameters, and the expanders $\G$, i.e., parameters that, if evaluated, could optimistically enlarge the safe set. For their formal definitions, see \cite{pmlr-v37-sui15} or \cref{sec:additional_defs}. During \textbf{LSE}, we evaluate the most uncertain parameter,  i.e., the parameter with the widest confidence interval, among the expanders and the maximizers:
\begin{equation}
    a_{n} =
    \underset{a \in \G \cup \M}{\arg\max} \; \underset{i \in \mathcal{I}}{\max } \; w_n(a,i),
    \label{eq:acq_lse}
\end{equation}
where $w_n(a,i) = u_n(a,i) - l_n(a,i)$. 

As a by-product of these experiments, \gosafeopt learns backup policies for all the states visited during these rollouts by leveraging the Markov property. 
Intuitively, for any state $x(t)$ visited when deploying a safe policy $a$ starting from $x_0$, we know that the sub-trajectory $\{x(\tau)\}_{\tau\geq t}$ is also safe because of Assumption \ref{ass:constraint}. Moreover, this sub-trajectory is safe regardless of how we reach $x(t)$ since the state is Markovian. Thus, $a$ is a valid backup policy for $x(t)$.  

This means we {\em learn about backup policies for multiple states during a single \gls{LSE} experiment}. To make them available during \textbf{GE}, we introduce the set of backups $\mathcal{B}_n \subseteq \mathcal{A} \times \mathcal{X}$. After running an experiment with policy $a$, we collect all the discrete state measurements in the rollout $\mathcal{R} = \underset{{k \in \mathbb{N}}}{\bigcup}\left\{a,x(k)\right\}$ and add it to the set of backups, $\mathcal{B}_{n+1} = \mathcal{B}_{n} \cup \mathcal{R}$ (see \cref{alg:LSE} line~\ref{lst:line:update_GP_lse}). 


We perform \gls{LSE} until the connected safe set is fully explored and the optimum within the safe set is discovered. Intuitively, this happens when we have learned our constraint and objective functions with high precision, i.e.,  when the uncertainty among the expanders and maximizers is less than $\epsilon$, and yet the safe set does not expand any further,
\begin{equation}
    \underset{a \in \mathcal{G}_{n-1} \cup \mathcal{M}_{n-1} }{\max} \underset{i \in \mathcal{I}}{\max } \; w_{n-1}(a,i) < \epsilon \text{ and } S_{n-1} = S_n.
    \label{eq:conv_lse}
\end{equation}
\looseness=-1
Note, \gosafeopt, like \safeopt, only explores the connected safe set in the parameter space and learns backup policies via the Markov property. 

\begin{algorithm}[t]
 \hbox{\textbf{Input}: Safe set $S$, set of backups $\mathcal{B}$, dataset $\mathcal{D}$}
\begin{algorithmic}[1]
\STATE Recommend parameter $a_n$ with  \cref{eq:acq_lse}
\STATE Collect $\mathcal{R}= \underset{{k \in \mathbb{N}}}{\bigcup}\left\{a_n,x(k)\right\}$ and $h(a_n,i)+\varepsilon_n$
\STATE $\mathcal{B} = \mathcal{B} \cup \mathcal{R}$, $\mathcal{D} =\mathcal{D} \cup \left\{a_n, h(a_n,i)+\varepsilon_n\right\}$ \label{lst:line:update_GP_lse}
\STATE Update sets $S$, $\mathcal{G}$, and $\mathcal{M}$ \COMMENT{\cref{eq:safeset}, \cref{sec:additional_defs} Definitions~\ref{def:expanders} and~\ref{def:maximizers}} \label{lst:line:update_S_lse}
\end{algorithmic}
 \hbox{\textbf{Return}: $S$, $\mathcal{B}$ ,$\mathcal{D}$}
\caption{Local Safe Exploration (\gls{LSE})}
\label{alg:LSE}
\end{algorithm}

\subsubsection{Global Exploration} \label{sec:global_exploration}
\begin{figure}[t]
   \begin{center}
\tikzset{every picture/.style={line width=0.75pt}} 
\begin{tikzpicture}[x=0.75pt,y=0.75pt,yscale=-1,xscale=1]

\draw  [color={rgb, 255:red, 208; green, 2; blue, 27 }  ,draw opacity=1 ] (142.43,94.14) .. controls (161.43,69.14) and (323.43,65.14) .. (269.43,93.14) .. controls (215.43,121.14) and (254.86,143.29) .. (241.43,177.14) .. controls (228,211) and (162,195) .. (142,165) .. controls (122,135) and (123.43,119.14) .. (142.43,94.14) -- cycle ;
\draw  [fill={rgb, 255:red, 74; green, 144; blue, 226 }  ,fill opacity=1 ] (172,99.43) .. controls (172,98.01) and (173.15,96.86) .. (174.57,96.86) .. controls (175.99,96.86) and (177.14,98.01) .. (177.14,99.43) .. controls (177.14,100.85) and (175.99,102) .. (174.57,102) .. controls (173.15,102) and (172,100.85) .. (172,99.43) -- cycle ;
\draw  [dash pattern={on 4.5pt off 4.5pt}] (136.14,99.43) .. controls (136.14,78.21) and (153.35,61) .. (174.57,61) .. controls (195.79,61) and (213,78.21) .. (213,99.43) .. controls (213,120.65) and (195.79,137.86) .. (174.57,137.86) .. controls (153.35,137.86) and (136.14,120.65) .. (136.14,99.43) -- cycle ;
\draw (161,105) node [anchor=north west][inner sep=0.75pt]   [align=left] {$x(t)$};
\draw (241,113) node [anchor=north west][inner sep=0.75pt]   [align=left] {\textcolor[rgb]{0.82,0.01,0.11}{$\min_{i \in \mathcal{I}_g}g_i(x) = 0$}};
\draw (128,195) node [anchor=north west][inner sep=0.75pt]   [align=left] {Trigger backup at $x(t)$};
\end{tikzpicture}
   \end{center}
   \caption{Illustration of the boundary condition. \capt{The backup policy is triggered at $x(t)$ if we cannot guarantee with high probability that all states in a ball around $x(t)$ are safe (see Assumption~\ref{ass:one_step_jump} and \cref{sec:global_exploration}). }}
   \label{boundary_condition_intuition}
\end{figure}
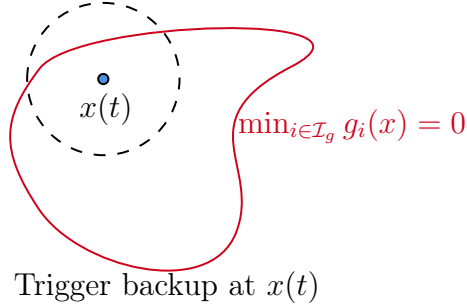

%
\looseness=-1
\gls{GE} aims at discovering new, disconnected safe regions. 
In particular, during a \gls{GE} step, we evaluate the most uncertain parameter, i.e., with the highest value for $\max_{i \in \mathcal{I}_g}w_n(a,i)$, outside of the safe set, $a$ $\in$ $\mathcal{A} \setminus S_n$. As this parameter is not in our safe set, it is not guaranteed to be safe. 
Therefore, we monitor the state during the experiment and trigger a backup policy, learned during \gls{LSE}, if we cannot guarantee staying in a safe region of the state space when continuing with the current choice of policy parameters (cf. \cref{boundary_condition_intuition}).

\looseness=-1
If a backup policy is triggered when evaluating the parameter $a$, we mark the experiment as failed. To avoid repeating the same experiment, we store $a$ and the state $x_{\textrm{Fail}}$ where we intervened in sets $\mathcal{E} \subset \mathcal{A}$ and $\mathcal{X}_{\textrm{Fail}} \subset \mathcal{X}$, respectively (see line~\ref{lst:line:fail_set_update} in \cref{alg:GE}). Thus, during \gls{GE}, we employ the following acquisition function
    \begin{equation}
        a_{n} =\underset{a \in \mathcal{A} \setminus \left(S_n \cup \mathcal{E}\right)}{\arg\max} \max_{i \in \mathcal{I}_g} w_n(a,i).
        \label{S3_acquisition}
    \end{equation}
This picks the most uncertain parameter, i.e., the parameter with the widest confidence interval, that is not provably safe but that has not been shown to trigger a backup policy.
If the experiment was run without triggering a backup, we know that $a$ is safe. Therefore, we add the observed values for $g_i$ and $f$ to the dataset and the 
rollout $\mathcal{R}$ collected during the experiment to our set of backups $\mathcal{B}_n$, i.e., $\mathcal{B}_{n+1} = \mathcal{B}_{n} \cup \mathcal{R}$.
Furthermore, we add the parameter $a$ to our safe set and update its lower bound, i.e., $l_n(a,i) = 0$, $\forall i \in \mathcal{I}_g$ (see lines~\ref{lst:line:update_GP_ge} and~\ref{lst:line:update_S_ge} in \cref{alg:GE}). Then, we switch to \gls{LSE} to explore the newly discovered safe area. Note, the lower bound is updated again before the \gls{LSE} step, i.e., $l_{n+1}(a,i) = \max\{l_n(a,i), \mu_{n}(a,i) - \beta^{\sfrac{1}{2}}_{n+1} \sigma_{n}(a,i)\}$ for all $i\in \mathcal{I}_g$ (see~\cref{alg:GSafeopt} line~\ref{lst:line:cn_update}).

If $\mathcal{A} \setminus \left(S_n \cup \mathcal{E}\right) = \emptyset$, there are no further safe areas we can discover and
\gls{GE} has converged. 

\begin{algorithm}[t]
 \hbox{\textbf{Input}: Safe set $S$, confidence intervals $C$, set of backups $\mathcal{B}$,} 
 \hbox{dataset $\mathcal{D}$, fail sets: $\mathcal{E}$, $\mathcal{X}_{\textrm{Fail}}$}
\begin{algorithmic}[1]
\STATE Recommend global parameter $a_n$  with \cref{S3_acquisition}
\STATE $a=a_n$, $x_{\textrm{Fail}} = \emptyset$, Boundary = False
\WHILE[Rollout policy]{Experiment not finished} \label{lst:line:experiment_start}
\STATE $x(k) = x_0 + \int\limits^{kT}_{t=0} z\left(x(t), \pi (x(t); a) \right) dt$
\IF[Not at boundary yet]{Not Boundary}
\STATE Boundary, $a^{*}_{s}$ = Boundary Condition($x(t),\mathcal{B}$)  \label{lst:line:start_BC}
\IF[Trigger backup policy]{Boundary}
\STATE $a= a^{*}_{s}$, $x_{\textrm{Fail}} = x(k)$ \label{lst:line:end_BC}
\STATE $\mathcal{E} = \mathcal{E}\cup \{a_n\}$, $\mathcal{X}_{\textrm{Fail}} = \mathcal{X}_{\textrm{Fail}}\cup \{x_{\textrm{Fail}}\}$ \COMMENT{update fail sets} \label{lst:line:fail_set_update}
\ENDIF 
\ENDIF
\ENDWHILE \label{lst:line:experiment_end}
\STATE Collect $\mathcal{R}= \underset{{k \in \mathbb{N}}}{\bigcup}\left\{a_n,x(k)\right\}$, and $h(a_n,i)+\varepsilon_n$
\IF[Successful global search]{Not Boundary}
\STATE $\mathcal{B} = \mathcal{B} \cup \mathcal{R}$ and $\mathcal{D} =\mathcal{D} \cup \left\{a_n,h(a_n,i)+\varepsilon_n\right\}$ \label{lst:line:update_GP_ge}
\STATE $S = S \cup a$, $C(a,i) = C(a,i) \cap [0, \infty]$ for all $i\in \mathcal{I}_g$.  \label{lst:line:update_S_ge}
\ENDIF
\end{algorithmic}
 \hbox{\textbf{Return}: $S$, $C$, $\mathcal{B}$, $\mathcal{D}$, $\mathcal{E}$, $\mathcal{X}_{\textrm{Fail}}$}
\caption{Global Exploration (\gls{GE})}
\label{alg:GE}
\end{algorithm}

\begin{algorithm}[t]
\hbox{\textbf{Input}: $x, \mathcal{B}_n$}
\begin{algorithmic}[1]
\IF{$\forall (a_s,x_s) \in \mathcal{B}_n, \exists i \in \mathcal{I}_g$, $l_n(a_s,i) - L_\mathrm{x} \left(\normAny{x - x_s} + \Xi\right) < 0$}
\STATE Boundary = True, Calculate $a^{*}_{s}$ (\cref{eq:backup_action})
\ELSE
\STATE Boundary = False, $a^*_s = \{\}$
\ENDIF
\end{algorithmic}
\hbox{\textbf{return}: Boundary, $a^{*}_{s}$}
\caption{Boundary Condition}
\label{BC_algo}
\end{algorithm}

\subsubsection{Boundary Condition.}
Throughout each \gls{GE} experiment, we monitor the state evolution, and, whenever a state measurement is received, we evaluate \textbf{\textit{online}} a boundary condition to determine whether a backup policy should be triggered. Ideally, it must \emph{(i)} guarantee safety, (\emph{ii}) be fast to evaluate even for high-dimensional dynamical systems, and (\emph{iii}) incorporate discrete-time measurements of the state. 
 To fulfill requirement \emph{(i)}, the boundary condition leverages Lipschitz continuity of the constraint.
In particular, when we are in $x\idxk$, we check if there is a point $(a_s,x_s)$ in our set of backups $\mathcal{B}_n$ such that $x_s$ is sufficiently close to $x\idxk$ to guarantee that $a_s$ can steer the system back to safety for any state we may reach in the next time step.

\heading{Boundary Condition:} 
    During iteration $n$, we trigger a backup policy at $x$ if there is no point in our set of backups $(a_s,x_s) \in \mathcal{B}_n$ such that $l_n(a_s,i) \geq L_\mathrm{x} \left(\normAny{x-x_s} + \Xi\right)$ for all $i \in \gls{conset}$. In this case, we use the backup parameter $a^{*}_{s}$ with the highest safety margin, that is
    \begin{equation}
        a^{*}_{s} =\underset{\left\{a_s \in \mathcal{A} | \exists x_s \in \mathcal{X}; (a_s,x_s) \in \mathcal{B}_n\right\}}{\max} \; \underset{i \in \mathcal{I}_g}{\min} \; l_n(a_s,i) - L_\mathrm{x}\normAny{x-x_s}. \label{eq:backup_action}
    \end{equation}
\looseness=-1
Since we already calculate $l_n(a_s,i)$ for all $i \in \mathcal{I}_g$ and $a_s \in S_n$ \textit{offline} to update the safe set (see \cref{eq:safeset}), we only need to evaluate $\normAny{x-x_s}$ \textit{online}, which is computationally tractable for most real-world systems (e.g., $\mathcal{O}(s)$ for the $2-$norm, where $s$ is the dimension of $\mathcal{X})$. 
Thereby, it satisfies requirement \emph{(ii)} and enables the application of our algorithm to complex systems with high sampling frequencies. 
The boundary condition is summarized in \cref{BC_algo}.


\heading{Updating Fail Sets.}
Parameters for which the boundary condition is triggered, i.e., parameters evaluated unsuccessfully during \gls{GE}, 
are added to the fail set $\mathcal{E}$. However, when \gls{LSE} is repeated after discovering a new region during \gls{GE}, we can learn new backup policies, which makes the boundary condition less restrictive. Hence, it may happen that a parameter $a$ for which
a backup policy was triggered during a previous \textbf{GE} step, i.e., $ a \in \mathcal{E}$, we would not trigger a backup policy after \textbf{LSE} step has converged in the new safe region. Thus, after learning new backup policies during \gls{LSE}, we re-evaluate the boundary condition (line~\ref{lst:line:recheck_Epsilon}),
and update $\mathcal{E}$ and $\mathcal{X}_{\textrm{Fail}}$ accordingly. 
These states may then be revisited during further \gls{GE} steps.

\looseness=-1
In summary, \gosafeopt involves two alternating stages, \gls{LSE} and \gls{GE}. \gls{LSE} steps are similar to \safeopt, nonetheless, they additionally leverage the Markov property of the system to learn backup policies. These backup policies are then used in \gls{GE} for global exploration. 
The only model-free safe exploration method that explores globally is \gosafe. However,  
it evaluates a completely different and expensive boundary condition, which relies on a safe set representation in the parameter and state space. This safe set is actively explored. Because of the active exploration, and expensive boundary condition, \gosafe becomes restricted to only systems with low-dimensional state spaces. 

\begin{remark}
\gosafeopt is devised for the episodic \gls{RL} setting where the initial state $x_0$ is fixed and known. In several applications, the initial state is not known apriori and instead sampled i.i.d.\ from a state distribution $\rho$. Our formulation can also be extended to this setting by treating the initial state as a context variable, c.f.~\cite{ContextualGPBO}. Moreover, to guarantee safety in this setting, Assumption~\ref{ass:safe_seed} has to be modified such that the parameters in the initial safe seed $S_0$, are safe for all initial states in the support of $\rho$, i.e., $x'_0 \in \text{supp}(\rho)$. 
Then, given a context/initial state $x'_0$, the acquisition function for \gls{LSE} or \gls{GE} is optimized for the context. This is similar to the contextual \safeopt algorithm~\cite{DBLP:journals/corr/BerkenkampKS16}. The boundary condition can also be extended to incorporate the context. Finally, for a continuous state space, $\text{supp}(\rho)$ can be discretized similarly to as in \gosafe. 
\end{remark}

\begin{algorithm}[t]
 \hbox{\textbf{Input}: Domain \gls{actionspace}, $k(\cdot,\cdot)$, $S_0$, $C_0$, $\mathcal{D}_0$, $\kappa$, $\eta$}
\begin{algorithmic}[1]
    \STATE Initialize \gls{GP} $h(a,i)$, $\mathcal{E}=\emptyset$, $\mathcal{X}_{\textrm{Fail}}=\emptyset$, $B_0 = \{(a,x_0) \mid a \in S_0\}$
    \WHILE{$S_n$ expanding or $\mathcal{A} \setminus \left(S_n \cup \mathcal{E}\right) \neq \emptyset$} \label{lst:line:conv_gosafeopt}
    \FOR[reevaluate fail sets]{$x \in \mathcal{X}_{\textrm{Fail}}$} \label{lst:line:recheck_Epsilon}
            \IF[\cref{BC_algo}]{Not \textbf{Boundary Condition}($x,\mathcal{B}_n$)}
                \STATE $\mathcal{E} = \mathcal{E}\setminus \{a\}$, $\mathcal{X}_{\textrm{Fail}} = \mathcal{X}_{\textrm{Fail}}\setminus \{x\}$ \COMMENT{Update fail sets} \label{lst:line:update_Epsilon}
            \ENDIF
\ENDFOR
        \STATE Update $C_n(a,i) \coloneqq [l_n(a,i), u_n(a,i)]$  $ \forall$ $a \in \mathcal{A}$, $i \in \mathcal{I}_g$ \label{lst:line:cn_update} \COMMENT{see \cref{safeopt_gosafe}}
        \IF[Perform \gls{LSE} (\cref{alg:LSE})]{\gls{LSE} not converged (\cref{eq:conv_lse})} \label{lst:line:lse_start}
            \STATE $S_{n+1}, \mathcal{B}_{n+1}, \mathcal{D}_{n+1} =$ \gls{LSE}($\mathcal{S}_{n},\mathcal{B}_{n},\mathcal{D}_{n}$)
        \ELSE[Perform \gls{GE} (\cref{alg:GE})]
            \STATE $S_{n+1}, C_{n+1}, \mathcal{B}_{n+1}, \mathcal{D}_{n+1}, \mathcal{E}, \mathcal{X}_{\textrm{Fail}} =$ \gls{GE}($\mathcal{S}_{n},C_n,\mathcal{B}_{n},\mathcal{D}_{n}, \mathcal{E}, \mathcal{X}_{\textrm{Fail}}$)
        \ENDIF
    \ENDWHILE
\end{algorithmic}
\hbox{\textbf{return}: $\underset{a \in S_n}{\arg\max}$ $l_n(a,0)$}
 \caption{\gosafeopt}
 \label{alg:GSafeopt}
\end{algorithm}
\subsection{Theoretical Results} \label{theoretical_guarantees}
This section provides safety (\cref{sec:safety_guarantees}) and optimality (\cref{sec:optimality_guarantees}) guarantees for \gosafeopt. 
\subsubsection{Safety Guarantees}\label{sec:safety_guarantees}
The main safety result for our algorithm is that \gosafeopt guarantees safety during all experiments.
\begin{restatable}[]{theorem}{safegosafeopt}
Under Assumptions \ref{ass:safe_seed} --  \ref{ass:constraint} and with $\beta_n$ as defined in \cite{DBLP:journals/corr/BerkenkampKS16}. \gosafeopt guarantees, for all $n\geq 0$ and any $\delta \in (0, 1)$, that experiments are safe as per Definition~\ref{def:safe_experiment} with probability at least $1-\delta$.
\label{thm:safety}
\end{restatable}
The proof of this theorem is provided in \cref{sec:BC_proof}.
Intuitively, we can analyze the safety of \gls{LSE} and \gls{GE} separately. For \gls{LSE}, we can leverage the results in \cite{DBLP:journals/corr/BerkenkampKS16}, which studies it extensively. Therefore, novel to our analysis is the safety of \gls{GE}. 
We show that while running experiments during \gls{GE}, we can guarantee that if our boundary condition triggers a backup, we are safe, and if a backup is not triggered, then the experiment is safe, i.e., we discovered a new safe parameter.

\subsubsection{Optimality Guarantees} \label{sec:optimality_guarantees}

Next, we analyze when \gosafeopt can find the safe global optimum $a^*$, which is the solution to \cref{eq:problem_statement}. 
During \gls{LSE}, we explore the connected safe region. For each safe region we explore, we can leverage the results from \cite{DBLP:journals/corr/BerkenkampKS16} to prove local optimality. Furthermore, due to \gls{GE}, we can discover disconnected safe regions and then repeat \gls{LSE} to explore them. 
To this end, we define when a parameter $a$ can be discovered by \gosafeopt (
 either during \gls{LSE} or during \gls{GE}). 
\begin{definition}
The parameter $a \in \mathcal{A}$ is discoverable by \gosafeopt at iteration $n$, if there exists a set $A \subseteq S_n$ such that $a \in \bar{R}^c_{\epsilon}(A)$. Here, $\bar{R}^c_{\epsilon}(A)$ is the largest safe set we can safely reach from $A$ (see \cref{eq:connected_set} in \cref{sec:proof_discoverable} or~\cite{DBLP:journals/corr/BerkenkampKS16, DBLP:gosafe}).
\label{def:discoverable_set}
\end{definition}
Next, we show that if the safe global optimum (solution of \cref{eq:problem_statement}) is discoverable as per Definition~\ref{def:discoverable_set}, then we can approximate it with $\epsilon$-precision.
\begin{restatable}[]{theorem}{optimality}
    Let $a^*$ be a safe global optimum. Further, let Assumptions~\ref{ass:safe_seed} --  \ref{ass:constraint} hold, $\beta_n$ be defined as in \cite{DBLP:journals/corr/BerkenkampKS16}. Assume there exists a finite integer $\tilde{n}\geq 0$ such that $a^*$ is discoverable at iteration $\tilde{n}$ (see~\cref{def:discoverable_set}). Then, for any $\epsilon>0$, and $\delta \in (0,1)$, there exists a finite integer $n^* \geq \tilde{n}$ such that with probability at least $1-\delta$,
    \begin{equation}
        f(\hat{a}_{n}) \geq f(a^*) - \epsilon, \quad  \forall n \geq n^*
    \end{equation}
    with $\hat{a}_{n} =\argmax_{a \in S_{n}} l_{n}(a,0)$. 
    \label{thm:optimality}
\end{restatable}
 In practice, \gosafeopt tends to find better controllers than \safeopt, which converges after \gls{LSE}. This is formalized in the following proposition.
\begin{proposition}
For \safeopt, $a^*$ is discoverable at iteration $n>0$, if and only if, it is discoverable at iteration $n=0$. 
\label{local_optimum_safeopt}
\end{proposition}
\Cref{local_optimum_safeopt} states that if the parameter $a^*$ does not lie in the largest safe set reachable from $S_0$, \safeopt will not find it. \gosafeopt does not suffer from the same restriction because of global exploration. 
In \cref{discoverable_proofs}, \cref{lemm:discoverable}
we provide additional conditions under which \gosafeopt can find the safe global optimum. The performance benefits for \gosafeopt are then empirically shown in~\cref{results}.

\remark The safety threshold $\delta$ is used to pick the designer's appetite for unsafe evaluations. For a large value of $\delta$, more parameters are available for sampling at each iteration. Accordingly, the method converges faster, however, while also allowing more unsafe evaluations, see~\cite{DBLP:journals/corr/BerkenkampKS16} for more detail. 
\subsection{Practical Modifications}
 In practice, we can further improve the sample and computational efficiency by introducing minor modifications. While they do not guarantee optimality, they yield good results for our evaluation in \cref{results}. Furthermore, all the proposed modifications \emph{do not affect the safety guarantees} of the method, and thus can be safely applied in practice.
\subsubsection{Fixing Iterations for Each Stage}
In \cref{alg:GSafeopt}, we perform a global search, i.e., \gls{GE}, after the convergence of \gls{LSE}. Nonetheless, it may be beneficial to run \gls{LSE} for a fixed amount of steps and then switch to \gls{GE}, before \textbf{LSE}'s convergence. This heuristic allows for the early discovery of disconnected safe regions, which may improve sample efficiency. 
Moreover, this allows ``jumping'' between different safe regions of the domain that, even though would be connected if we ran the current \textbf{LSE} to convergence, are currently disconnected. 
To this end, we apply the following heuristic scheme: (\emph{i}) run \gls{LSE} for $n_{\textrm{LSE}}$ steps, (\emph{ii}) run \gls{GE} for $n_{\textrm{GE}}$ steps or until we have discovered new safe parameters, and (\emph{iii}) if \gls{GE} discovers a new region, return to (\emph{i}). Else, return to (\emph{i}) after \gls{GE} completion, but with reduced $n_{\textrm{LSE}}$.
Note, the proposed scheme still retains optimality because we do not restrict the total number of iterations with the system. However, in practice, we additionally impose an upper bound on the interactions, and therefore  $n_{\textrm{LSE}}$, and $n_{\textrm{GE}}$ influence the budget of global and local exploration, this affects optimality (c.f., \cref{sec:ablation_study}).

\subsubsection{Updated Boundary Condition}\label{sec:UBC}
If required, the boundary condition can be further modified to reduce computation time by considering only a subset of the states collected from experiments. The updated boundary condition reduces the online computation time at the expense of a more conservative boundary condition. Due to this conservatism, we lose our optimality guarantees. In practice, however, we still achieve good results (see \cref{results}).
\begin{definition}
Consider $\eta_l \in \R$ and $\eta_u \in \R$ such that $\eta_l < \eta_u$. The interior set $\Omega_{I, n}$ and marginal set $\Omega_{M, n}$ are defined as
\begin{align*}
\Omega_{I, n} &= \{x_s \in \gls{statespace} \mid (a,x_s) \in \mathcal{B}_n: \forall i \in \gls{conset},  l_n\left(a,i\right)\geq \eta_u\} \\
    \Omega_{M, n} &= \{x_s \in \gls{statespace}\mid (a,x_s) \in \mathcal{B}_n: \forall i \in \gls{conset},  \eta_l \leq l_n\left(a,i\right)<\eta_u\}.
\end{align*}
\end{definition}
The interior set contains the points in our set of backups $\mathcal{B}_n$ that are safe with high tolerance $\eta_u$, whereas the points in the marginal set are safe with a smaller tolerance $\eta_l$.
We use those sets for the updated boundary condition. 

\heading{Updated Boundary Condition}:
Consider $d_l \in \R$ and $d_u \in \R$ such that $d_l < d_u$.
We trigger a backup policy at $x$ if there is not a point $x_s \in \Omega_{I,n}$ such that $\normAny{x-x_s} \leq d_u$ or there is not a point $x'_s \in \Omega_{M,n}$ such that $\normAny{x-x'_s} \leq d_l$. In this case, we use the backup parameter $a^*_s$
\begin{equation}
        a^{*}_{s} =\underset{\left\{a_s \in \mathcal{A} | (a_s,x_s) \in \mathcal{B}_n\right\}}{\max} \; l_n(a_s,i); \quad  \text{with }x_s = \underset{x' \in \Omega_{I, n} \cup\Omega_{M, n}}{\min} \; \normAny{x-x'}. \label{eq:backup_action_updated}
    \end{equation}
Intuitively,  we define distance tolerances $d_u$, and $d_l$ for points in $\mathcal{B}_n$ based on their safety tolerances $\eta_u, \eta_l$. As for \cref{thm:safety}, we can derive appropriate values for $\eta_u, d_u$, respectively $\eta_l,d_l$ to guarantee safety.
\section{Evaluation}\label{results}
\begin{figure}[thbp]
   \centering
   \includegraphics[width=0.45\textwidth]{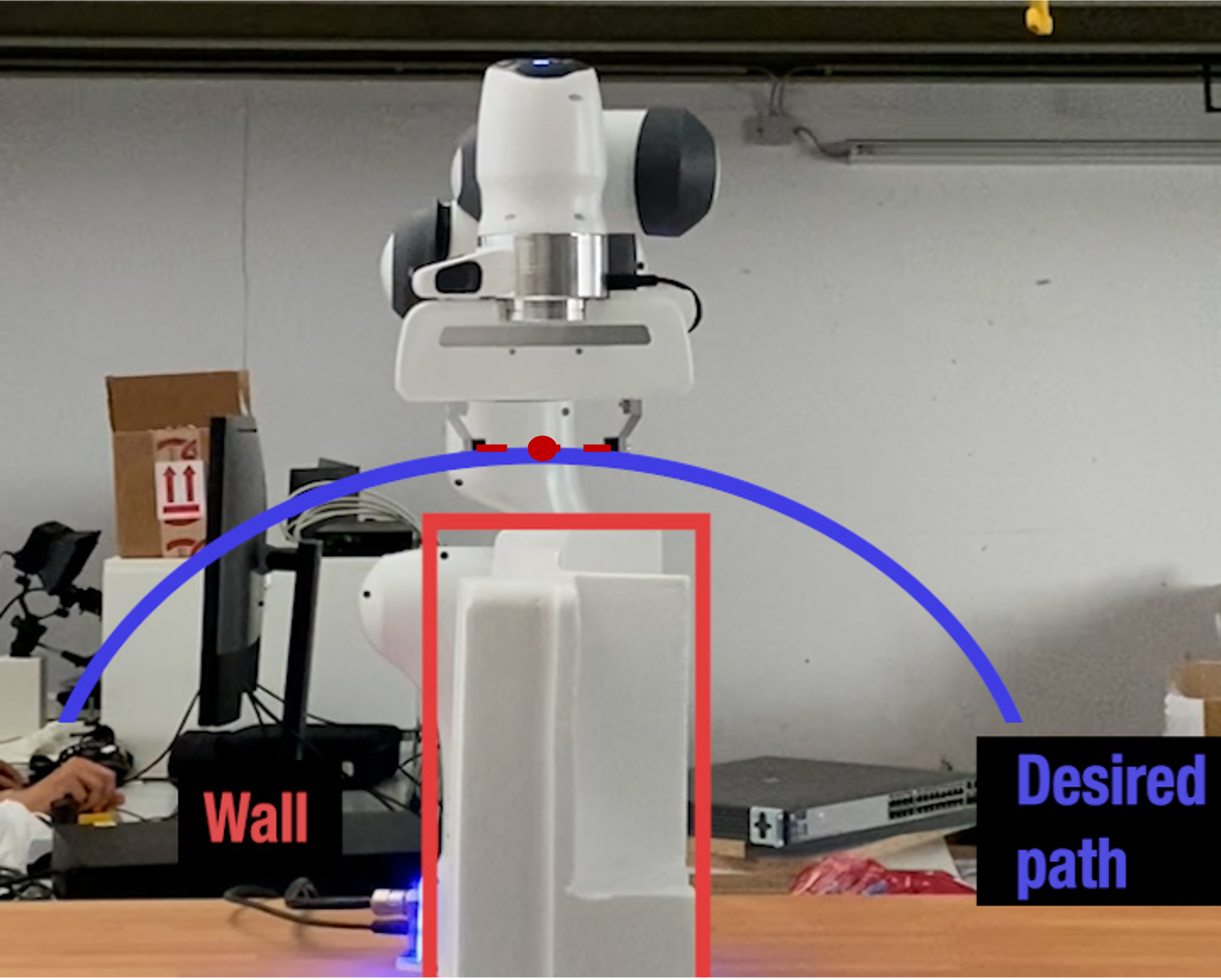}
   \captionsetup{width=.9\textwidth}
    \captionof{figure}{Setup for our evaluation in \cref{results}. \capt{We consider a safety-critical path following problem where deviations from the desired path (blue) could cause the robot to hit the wall (red box) and incur damage.}}
    \label{fig:hardware_path}
\end{figure}

We evaluate \gosafeopt in simulated and real experiments on a Franka Emika Panda seven \gls{DOF} robot arm\footnote{A video of our hardware experiments and link to code are available: 
\url{https://sukhijab.github.io/GoSafeOpt/main_project.html}} (see \cref{fig:franka_panda} and~\ref{fig:hardware_path}). 
The objective of our experiments is to demonstrate that \gosafeopt \emph{(i)}  can be applied to systems with high dimensional state spaces,  \emph{(ii)} is successful in safely tuning control parameters in common robotic tasks such as path following with manipulators, and \emph{(iii)} is superior to the existing state-of-the-art method, \safeopt, for safe control parameter tuning of real-world robotic systems.
\vspace{1em}

Accordingly, in our results, we show that \gosafeopt can scale to high dimensional systems, jump to disconnected safe regions while guaranteeing safety, and is directly applicable to hardware tasks with high sampling frequencies. 
In this work, we do not consider very high dimensional parameter spaces, which are in themselves challenging to tackle for methods such as \safeopt. Methods in ~\cite{DUIVENVOORDEN201711800} and~\cite{DBLP:journals/corr/abs-1902-03229} alleviate this challenge and can be integrated with our algorithm easily. Thus, we concentrate on the novelty of our method, which is its globally safe parameter exploration, unlike \safeopt, and scalability to high-dimensional state spaces compared to \gosafe. Specifically, the state space of systems we consider in this section is too large for \gosafe and it cannot be applied to any of our problems.

Details on the objective and constraint functions are provided in \cref{exp_info}. The hyperparameters of our experiments are listed in \cref{hyperparams}. 

In all experiments with the robot arm, we solely control the position and velocity of the end-effector. To this end, we consider an operational space impedance controller~\cite{robotics_handbook} with impedance gain $K$ (see \cref{exp_info}). The state space for our problem is six-dimensional. This is prohibitively large for \gosafe (struggles with state space greater than three~\cite{DBLP:journals/corr/abs-1902-03229}). Therefore, we compare our method with \safeopt. 

\looseness=-1
Impedance controllers for manipulators are usually tuned manually. This is often a tedious and time-consuming process. Accordingly, we show in our results that \gosafeopt can be used to automate this tuning safely. 
\subsection{Simulation Results}\label{8D_task}
We first evaluate \gosafeopt in a simulation environment based on the Mujoco physics engine~\cite{mujoco}\footnote{The URDFs and meshes are taken from~\url{https://github.com/StanfordASL/PandaRobot.jl}}. For this we consider two distinct tasks, (\emph{i}) reaching a desired position, and (\emph{ii}) path following. We determine the impedance gain through an approximate model of the system and perform feedback linearization~\cite{robotics_handbook}. For the resulting linear system, we design 
a \gls{lqr}~\cite{bertsekas} with quadratic costs that are parameterized by matrices $Q \in \R^{n\times n}$ and $R \in \R^{p \times p}$. Since the model is inaccurate, the feedback linearization will not cancel all nonlinearities and the \gls{lqr} will not be optimal.
Thus, our goal is to tune the cost matrices $Q$ and $R$ to compensate for the model mismatch. This approach is similar to~\cite{DBLP:journals/corr/MarcoHBST16}. We evaluate our methods over twenty independent runs of $200$ iterations. 

\subsubsection{Task 1: Reaching a Desired Position}
\begin{figure}
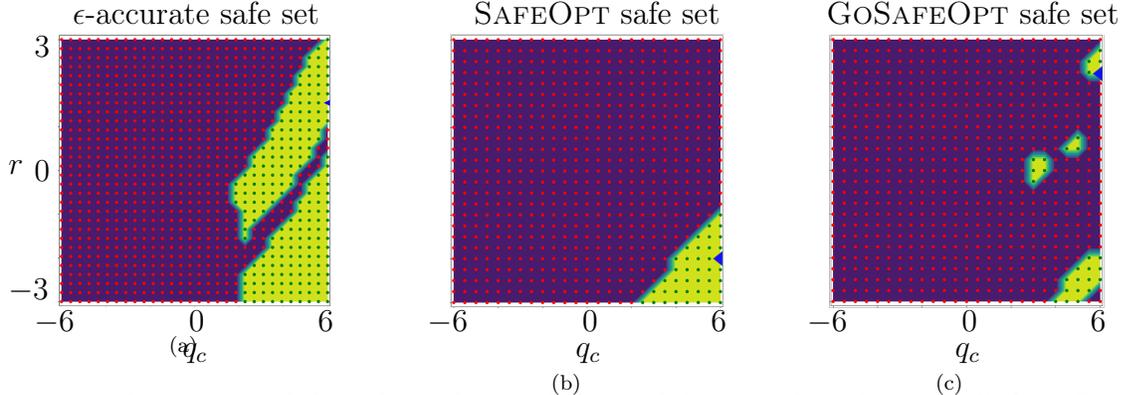

   \begin{subfigure}[t]{0.3\textwidth}
    \begin{center}
        \input{safset}
      \vspace*{-1.5em}
          \caption{}
                \end{center}
    \end{subfigure}
    \begin{subfigure}[t]{0.3\textwidth}
    \begin{center}
   \input{safeoptsafeset}
        \vspace*{-1.5em}
        \hspace{1cm}
    \caption{}
        \end{center}
    \end{subfigure}
    \begin{subfigure}[t]{0.3\textwidth}
    \begin{center}
        \input{gosafeoptsafaset}
    \vspace*{-1.5em}
      \hspace{1cm}
    \caption{}
        \end{center}
    \end{subfigure}
    \caption{Comparison of the safe set for simulation task between \safeopt and \gosafeopt after $200$ iterations.\capt{ The yellow regions represent the safe sets. 
    In each figure, the optimum is represented by a blue triangle. 
    }}
    \label{safesets_sim}
    \end{figure}

\looseness=-1
We select a target $x_{\mathrm{des}} \in \R^3$ for the robot. For this task, we parameterize the matrices $Q$ and $R$ by two parameters $(q_c, r)$ $\in [2,6] \times [-3,3]$, that trade-off accurate tracking, i.e., large $q_c$, and small inputs, i.e., large $r$. We choose the objective function to encourage reaching the target as fast as possible while penalizing large end-effector velocities and control actions (see 
\cref{FR_sim} for details). Thus in total, we have an eight-dimensional task (six-dimensional state space and two-dimensional parameter space). 
For analysis purposes, we run a simple grid search, that we could not run outside of simulation, to get an estimate of the safe set and the global optimum. \cref{safesets_sim} depicts the $\epsilon$-precise ($\epsilon=0.1$) safe set observed via grid search. From the figure, we observe that 
there is a disconnected safe region. 

\heading{Evaluation:}
  \cref{safesets_sim} depicts the safe sets of \safeopt and \gosafeopt after $200$ learning iterations.
We see that  
 \safeopt cannot discover the disconnected safe region and hence is stuck at a local optimum.
 On the other hand, \gosafeopt discovers the disconnected regions and can  jump within connected safe sets. 
The learning curve of the two methods is depicted in \cref{sfig:software_rewards}. Our method performs considerably better than \safeopt. The optimum found by our method is $0.007$ (less than $\epsilon=0.1$) close to the optimum found via the grid search. \safeopt cannot significantly improve over the initial policy. This is because the initial safe seed $S_0$ already contains a near-optimal policy from the connected region \safeopt explores, i.e., $\max_{a \in S_0} f(a) \approx \max_{a \in \Bar{R}^c_{\epsilon}(S_0)} f(a)$.
Lastly, our method also achieves comparable safety to \safeopt (on average $99.9\%$ compared to $100\%$). We encounter the failures during \gls{LSE}, 
which corresponds to \safeopt, one could also expect similar behavior from \safeopt if it were initialized in the upper region. 

\vspace{-0.5em}
\remark We can increase $\beta_n$ to encourage conservatism and avoid all unsafe evaluations. However, this also influences the algorithm's convergence rate. Hence, in practice, based on the task and appetite for unsafe evaluations $\beta_n$ has to be selected.

\subsubsection{Task 2: Path Following Task} 
For this experiment we define a parameterized path for the robot arm to follow  $x_{\mathrm{d}}(\rho(t))$. Here, we define $\rho(t)$ as a state to indicate progress along the trajectory, i.e.,  $x_{\mathrm{d}}(0) = x_0$, $x_{\mathrm{d}}(1) = x_{\mathrm{des}}$. The evolution of $\rho(t)\in [0,1]$ is controlled by a parameter $a_{\rho} \in [0, 1]$, that is, $\rho(t)=\min\{t(a_{\rho}\left(\sfrac{1}{100}-\sfrac{1}{500}\right) + \sfrac{1}{500}),1\}$. The objective is to find optimal control parameters for $Q, R$, and $a_{\rho}$ such that we progress on $x_{\mathrm{d}}(\cdot)$ as fast as possible while ensuring that $\normgeneral{x-x_{\mathrm{d}}\big(\rho(t)\big) }{2} \leq \zeta$. 
In this example, we model $Q, R$ using three parameters, $q_c,r,\kappa_d$, where $\kappa_d \in [0, 1]$ is used to weigh the velocity cost with respect to the positional cost of our state in the $Q$ matrix (c.f.~\cref{FR_sim}). Together with $a_{\rho}$ as a parameter, this task is eleven-dimensional, with seven states (including $\rho$) and four parameters.  This problem incorporates a challenging trade-off between fast trajectories and high-tracking performance. 
We compare it to \safeoptswarm~\cite{DUIVENVOORDEN201711800}, a scalable version of \safeopt for larger parameter spaces that use adaptive discretization. 
The results are presented in~\cref{sfig:software_rewards_11d}. Our results again show that \gosafeopt performs considerably better than \safeopt, specifically \safeoptswarm. Furthermore, both \safeopt and \gosafeopt give $100\%$ safety over all 20 runs. We also compare our method with expected improvement with constraints (EIC)~\cite{Gelbart} in \cref{sfig:eic_comparison}. EIC discourages potentially unsafe regions but allows for unsafe evaluations. Our results show that EIC, and \gosafeopt attain similar performance. However, EIC has considerably more unsafe evaluations (on average greater than fifteen) than \gosafeopt, which has none. 

\subsection{Hardware Results}
While the simulation results already showcased the general applicability of \gosafeopt to high dimensional systems and its ability to discover disconnected safe regions, we now demonstrate that it can also safely optimize policies on real-world systems.

\heading{Control Task:} 
We consider a path following task (see the experimental setup in \cref{fig:hardware_path}), and model the impedance gain $K$ as
\begin{equation*}
    K =\textrm{diag}\left(K_x,K_y,K_z,2\sqrt{K_x},2\sqrt{K_y},2\sqrt{K_z}\right),
\end{equation*}
\looseness=-1
where $K_x\!=\! \alpha_x K_{r,x}$ with $K_{r,x}\! >\! 0$ a reference value used for Franka's impedance controller and $\alpha_x \in [0, 1.2]$ the parameter we would like to tune (same for $y,z$). Accordingly, $\alpha_{x,y,z}=1$ corresponds to the impedance controller provided by the manufacturer. The parameter space we consider for this task is $[0,1.2]^3$. 
We require the controller to follow the known desired path while avoiding the wall depicted in \cref{fig:hardware_path}. 

\heading{Optimization Problem:}  We choose our objective function to encourage tracking the desired path as accurately as possible and impose a constraint on the end-effector's distance from the wall (see \cref{Hw_results} for more details).  
We receive a measurement of the state at \SI{250}{\hertz} and evaluate the boundary condition during \gls{GE} at \SI{100}{\hertz}. 

\begin{figure}[t]
    \begin{center}
        \input{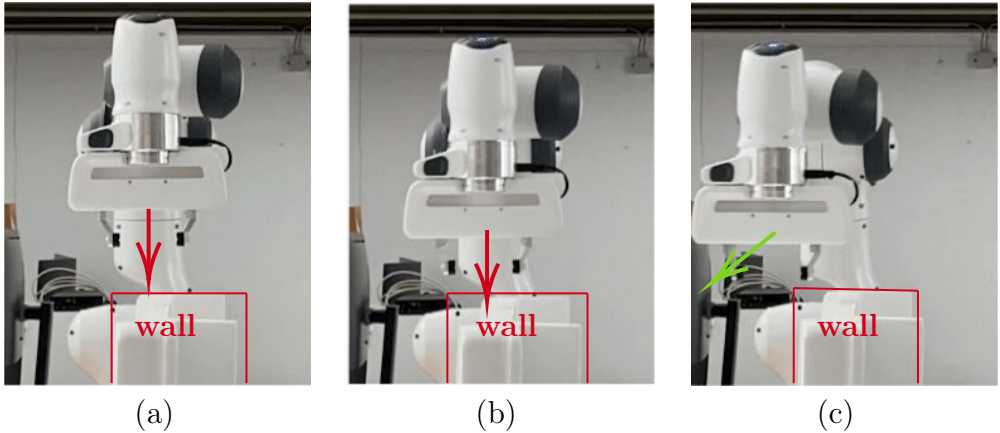}
    \end{center}
    \vspace*{-1em}
    \caption{Illustration of triggering the backup policy during \gls{GE}. \capt{During the global search, the policy directs the robot towards the wall in (a) and (b). A backup policy is automatically triggered by our boundary condition, once the robot gets too close to the wall. The backup policy directs the robot away from the wall (see green arrow in (c)).}}
      \label{fig:hardware_backup}
\end{figure}

\begin{figure}[!ht]
\begin{minipage}{0.5\textwidth}
    \begin{center}
        \includegraphics[width=0.9\textwidth]{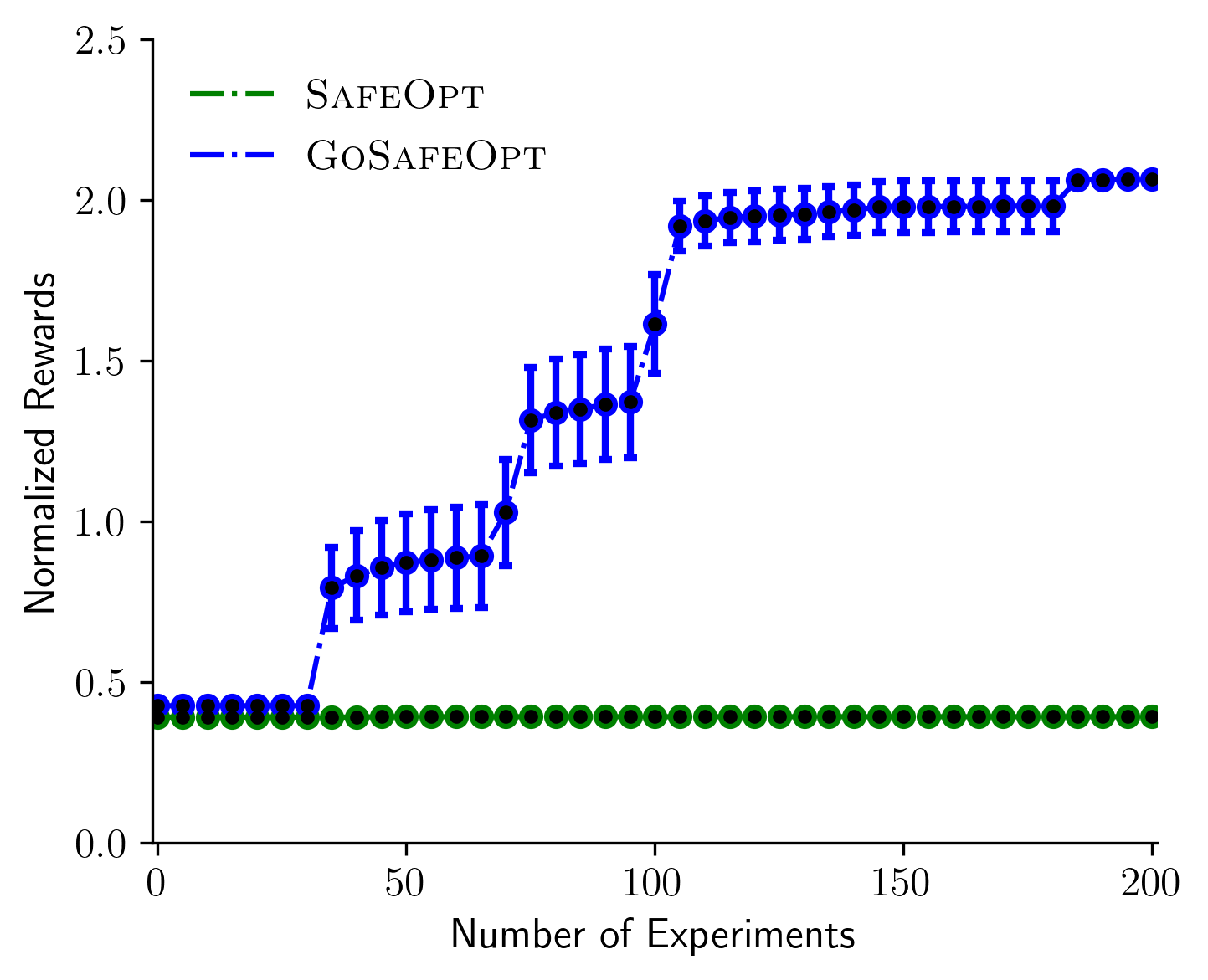}
        \captionsetup{width=.9\textwidth}
    \caption{Mean normalized objective with standard error for \safeopt and \gosafeopt for the eight-dimensional simulation task (20 runs).} 
    \label{sfig:software_rewards}
    \end{center}
    \end{minipage}
        \begin{minipage}{0.5\textwidth}
    \begin{center}
        \includegraphics[width=0.9\textwidth]{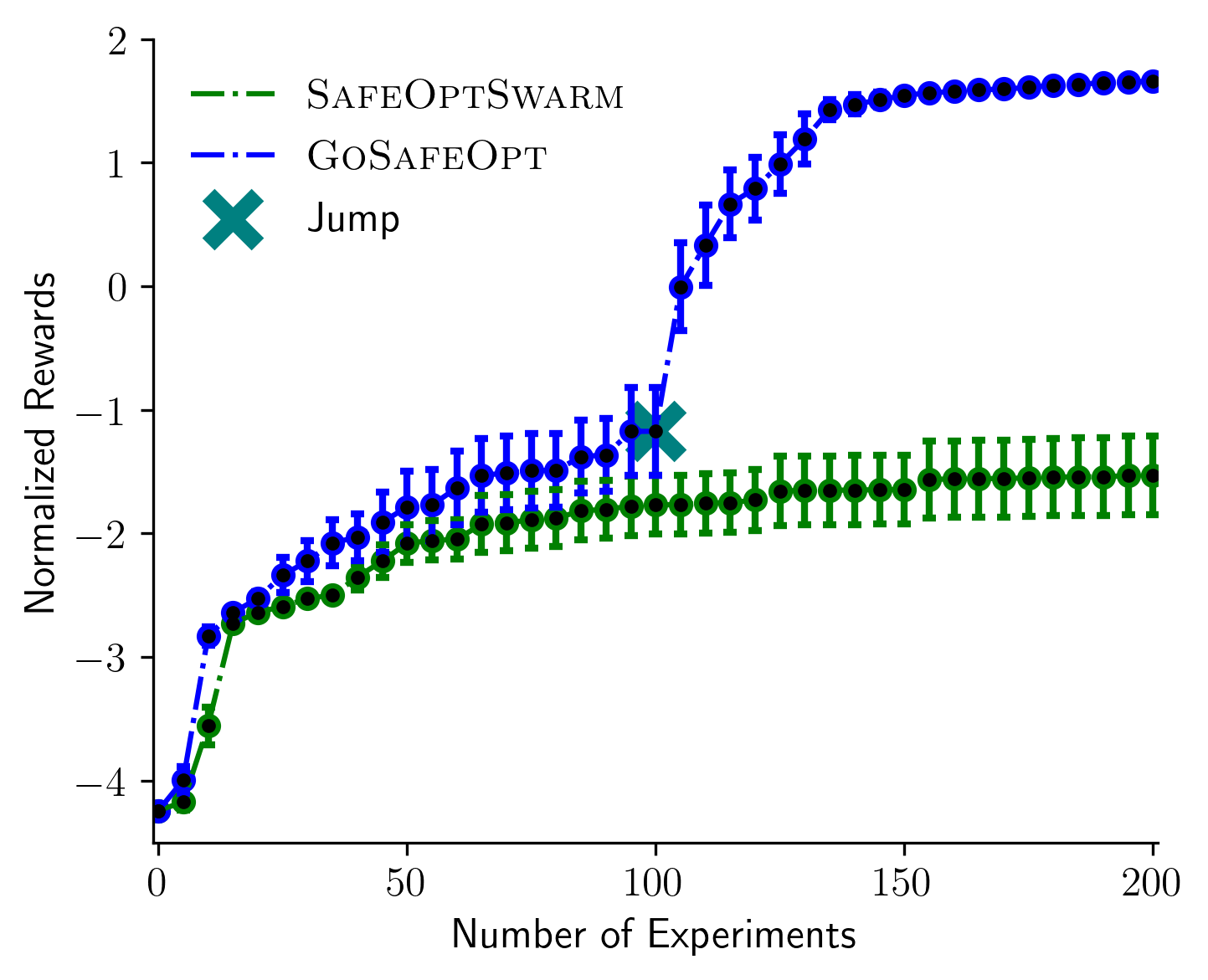}
        \captionsetup{width=.9\textwidth}
    \caption{Mean normalized objective with standard error for \safeopt and \gosafeopt for the eleven-dimensional simulation task (20 runs).}
    \label{sfig:software_rewards_11d}
    \end{center}
    \end{minipage}
    \begin{minipage}{0.5\textwidth}
    \begin{center}
        \includegraphics[width=0.9\textwidth]{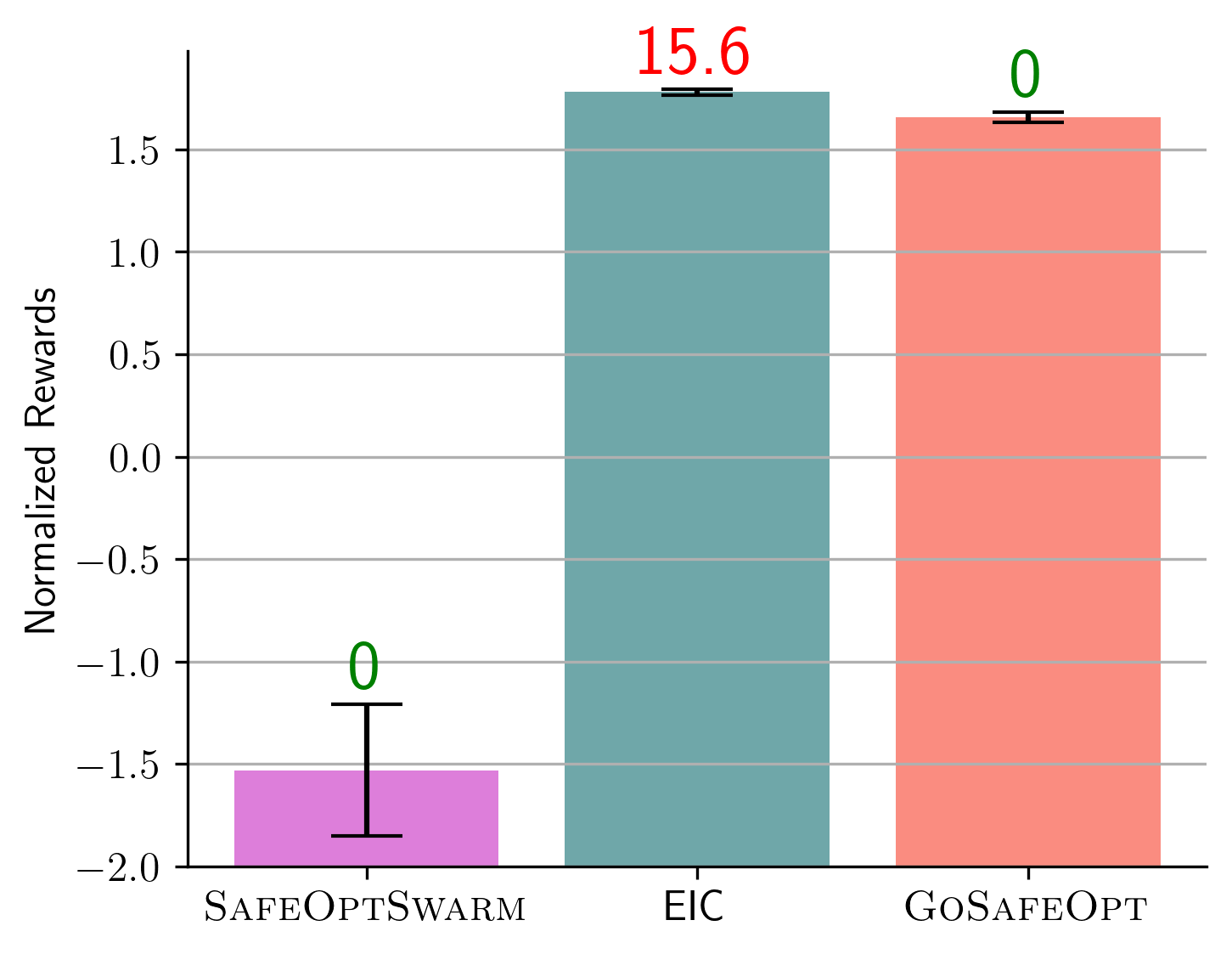}
        \captionsetup{width=.9\textwidth}
    \caption{Comparison of the normalized rewards with standard error and number of unsafe evaluations (numbers on top of the bars) between \safeoptswarm, EIC, and \gosafeopt for the eleven-dimensional simulation task (20 runs).}
    \label{sfig:eic_comparison}
    \end{center}
    \end{minipage}
            \begin{minipage}{0.5\textwidth}
    \begin{center}
        \includegraphics[width=0.9\textwidth]{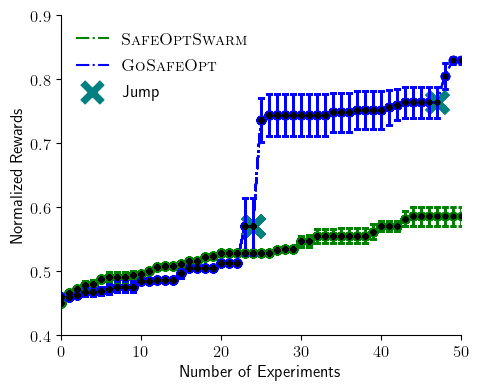}
    \caption{
    Mean normalized objective with standard error for \safeoptswarm and \gosafeopt for the hardware task (3 runs). \capt{The approximate location of the jump during \gls{GE} is visible and indicated with a cyan cross.}}
    \label{sfig:hardware_rewards}
    \end{center}
    \end{minipage}
    \end{figure}

\heading{Evaluation:}
The parameter space for this task is three-dimensional. Therefore, we compare our method to \safeoptswarm \cite{DUIVENVOORDEN201711800} and run only 50 iterations for each algorithm in three independent runs. 
We choose $a_0 = (0.6,0.6,0.6)$ as our initial policy. During our experiments, both \gosafeopt and \safeoptswarm provide $100 \%$ safety in all three runs. 
For \gosafeopt, safety during \gls{GE} is preserved by triggering a backup policy if required. One such instance is shown in~\cref{fig:hardware_backup}. 
We see in \cref{sfig:hardware_rewards} that \gosafeopt performs considerably better than \safeoptswarm. In particular, even if we cannot prove the existence of disconnected safe regions for this task, \gosafeopt still finds a better policy due to \gls{GE}. Interestingly, the optimal value suggested by \gosafeopt for both $\alpha_{x}$, and $\alpha_{y}$ is $1.2$. Therefore, in the direction of our path, \gosafeopt suggests aggressive controls to reduce tracking error. Moreover, the controller suggested by \gosafeopt is more aggressive than the manufacturers' reference controller ($\alpha_{x} = 1.0$, $\alpha_{y}=1.0$), and tracks the trajectory better.
\subsubsection{Choosing Hyperparameters}
\label{choosing_hyperparams}
\gosafeopt, like many safe exploration \gls{BO} algorithms such as \safeopt and \gosafe, makes assumptions on prior knowledge of the system (see~\cref{sec:accumption}). These assumptions are crucial for theoretical guarantees. In practice, they are hard to verify. Yet, safe exploration \gls{BO} methods have been successfully and safely applied to a large breath of applications~\cite{sui2018stagewise, racecarSafeopt, siemensSafeOpt, GooseApplied, CooperSafeOpt}. In our case, we leverage the available simulator to obtain a range for the hyperparameters: kernel parameters, $\beta_n$, and distance metric for the boundary condition. Lastly, with $\beta_n$ fixed, we fine-tune the remaining parameters by performing
controlled safe experiments with the hardware. Even though this approach gives good results, recent work from~\cite{rothfuss2022meta} investigates the hyperparameter selection problem for safe \gls{BO} more systematically. In general, there are
a few other works which investigate the gap between theory and practice~\cite{tight_bounds, hyperParamFelix}. 
Nonetheless, given the potential of these algorithms for reliable and safe \gls{AI}, we acknowledge
that future research on bridging this gap is needed.


\section{Conclusion}
\looseness=-1
This work proposes \gosafeopt, a novel model-free learning algorithm for global safe optimization of policies for complex dynamical systems with high-dimensional state spaces.
%
We provide for \gosafeopt high probability safety guarantees and show that it provably performs better than \safeopt, a state-of-the-art model-free safe exploration algorithm. 
We demonstrate the superiority of our algorithm over \safeopt empirically through our experiments. 
%
\gosafeopt can handle more complex and realistic dynamical systems compared to existing model-free learning methods for safe global exploration, such as \gosafe. This is due to a combination of an efficient passive discovery of backup policies  that leverages the Markov property of the system and a novel and efficient boundary condition to detect when to trigger a backup policy.
Future extensions could design hybrid algorithms that leverage the Markov property \textit{and} actively explore the state space. Moreover, GoSafeOpt is designed for efficient and safe controller tuning. We believe it can be applied to other dynamical systems, e.g., in legged robotics, where controller parameter tuning is a crucial component~\cite{leggedrobotics}. 
\section*{Acknowledgements}

We would like to thank Kyrylo Sovailo for helping us with the hardware experiments on the Franka Emika Panda arm and Alonso Marco for insightful discussions and for providing the EIC code. 
Furthermore, we would also like to thank Christian Fiedler, Pierre-François Massiani, and Steve Heim for their feedback on this work. 

This project has received funding from the Federal Ministry of Education and Research (BMBF) and
the Ministry of Culture and Science of the German State of North Rhine-Westphalia
(MKW) under the Excellence Strategy of the Federal Government and the Länder, the European Research Council (ERC) under the European Union's Horizon 2020 research and innovation program grant agreement No 815943, the Swiss National Science Foundation under NCCR Automation, grant agreement 51NF40 180545, and the Microsoft Swiss Joint Research Center.
\Urlmuskip=0mu plus 1mu\relax
\bibliographystyle{elsarticle-num.bst}
\bibliography{references}

\begin{thebibliography}{10}
\expandafter\ifx\csname url\endcsname\relax
  \def\url#1{\texttt{#1}}\fi
\expandafter\ifx\csname urlprefix\endcsname\relax\def\urlprefix{URL }\fi
\expandafter\ifx\csname href\endcsname\relax
  \def\href#1#2{#2} \def\path#1{#1}\fi

\bibitem{Sutton1998}
R.~S. Sutton, A.~G. Barto, Reinforcement Learning: An Introduction, 2nd
  Edition, The MIT Press, 2018.

\bibitem{levine2016end}
S.~Levine, C.~Finn, T.~Darrell, P.~Abbeel, End-to-end training of deep
  visuomotor policies, The Journal of Machine Learning Research 17~(1) (2016)
  1334--1373.

\bibitem{peters2008reinforcement}
J.~Peters, S.~Schaal, Reinforcement learning of motor skills with policy
  gradients, Neural networks 21~(4) (2008) 682--697.

\bibitem{lillicrap2015continuous}
T.~P. Lillicrap, J.~J. Hunt, A.~Pritzel, N.~Heess, T.~Erez, Y.~Tassa,
  D.~Silver, D.~Wierstra, Continuous control with deep reinforcement learning,
  arXiv preprint arXiv:1509.02971 (2015).

\bibitem{kober2013reinforcement}
J.~Kober, J.~A. Bagnell, J.~Peters, Reinforcement learning in robotics: A
  survey, The International Journal of Robotics Research 32~(11) (2013)
  1238--1274.

\bibitem{schaal2010learning}
S.~Schaal, C.~G. Atkeson, Learning control in robotics, IEEE Robotics \&
  Automation Magazine 17~(2) (2010) 20--29.

\bibitem{mockus1978application}
J.~Mockus, V.~Tiesis, A.~Zilinskas, The application of {B}ayesian methods for
  seeking the extremum, Towards Global Optimization 2~(117-129) (1978) 2.

\bibitem{Calandra2016}
R.~Calandra, A.~Seyfarth, J.~Peters, M.~Deisenroth, Bayesian optimization for
  learning gaits under uncertainty, Annals of Mathematics and Artificial
  Intelligence 76 (2016) 5--23.

\bibitem{DBLP:journals/corr/MarcoHBST16}
A.~Marco, P.~Hennig, J.~Bohg, S.~Schaal, S.~Trimpe, Automatic {LQR} tuning
  based on {G}aussian process global optimization, in: IEEE International
  Conference on Robotics and Automation, 2016, pp. 270--277.

\bibitem{DBLP:journals/corr/AntonovaRA17}
R.~Antonova, A.~Rai, C.~G. Atkeson, Deep kernels for optimizing locomotion
  controllers, in: Conference on Robot Learning, 2017, pp. 47--56.

\bibitem{DBLP:journals/corr/abs-1910-13399}
M.~Turchetta, A.~Krause, S.~Trimpe, Robust model-free reinforcement learning
  with multi-objective {B}ayesian optimization, in: IEEE International
  Conference on Robotics and Automation, 2020, pp. 10702--10708.

\bibitem{Gelbart}
M.~Gelbart, J.~Snoek, R.~Adams, Bayesian optimization with unknown constraints,
  Conference on Uncertainty in Artificial Intelligence (2014) 250–259.

\bibitem{hernandez2016general}
J.~M. Hern\'{a}ndez-Lobato, M.~A. Gelbart, R.~P. Adams, M.~W. Hoffman,
  Z.~Ghahramani, A general framework for constrained {B}ayesian optimization
  using information-based search, The Journal of Machine Learning Research
  17~(1) (2016) 5549--5601.

\bibitem{classRegress}
A.~Marco, D.~Baumann, M.~Khadiv, P.~Hennig, L.~Righetti, S.~Trimpe, Robot
  learning with crash constraints, IEEE Robotics and Automation Letters 6~(2)
  (2021) 1439--1446.

\bibitem{heim2020learnable}
S.~Heim, A.~Rohr, S.~Trimpe, A.~Badri-Spr{\"o}witz, A learnable safety measure,
  in: Conference on Robot Learning, 2020, pp. 627--639.

\bibitem{pmlr-v37-sui15}
Y.~Sui, A.~Gotovos, J.~Burdick, A.~Krause, Safe exploration for optimization
  with {G}aussian processes, in: International Conference on Machine Learning,
  2015, pp. 997--1005.

\bibitem{DBLP:journals/corr/BerkenkampSK15}
F.~Berkenkamp, A.~P. Schoellig, A.~Krause, Safe controller optimization for
  quadrotors with {G}aussian processes, in: IEEE International Conference on
  Robotics and Automation, 2016, pp. 491--496.

\bibitem{DBLP:journals/corr/BerkenkampKS16}
F.~Berkenkamp, A.~Krause, A.~P. Schoellig, Bayesian optimization with safety
  constraints: safe and automatic parameter tuning in robotics, Machine
  Learning (2021).

\bibitem{koenig2021safe}
C.~König, M.~Turchetta, J.~Lygeros, A.~Rupenyan, A.~Krause, Safe and efficient
  model-free adaptive control via {B}ayesian optimization, arXiv preprint
  arXiv:2101.07825 (2021).
\newblock \href {http://arxiv.org/abs/2101.07825} {\path{arXiv:2101.07825}}.

\bibitem{GRYAZINA200613}
E.~N. Gryazina, B.~T. Polyak, Stability regions in the parameter space:
  {D}-decomposition revisited, Automatica 42~(1) (2006) 13--26.

\bibitem{DBLP:gosafe}
D.~Baumann, A.~Marco, M.~Turchetta, S.~Trimpe, Go{S}afe: Globally optimal safe
  robot learning, in: IEEE International Conference on Robotics and Automation,
  2021, pp. 4452--4458, {P}roofs in extended online version: ar{X}iv
  2105.13281.

\bibitem{DBLP:journals/corr/abs-1902-03229}
J.~Kirschner, M.~Mutny, N.~Hiller, R.~Ischebeck, A.~Krause, Adaptive and safe
  {B}ayesian optimization in high dimensions via one-dimensional subspaces, in:
  International Conference on Machine Learning, 2019, pp. 3429--3438.

\bibitem{sui2018stagewise}
Y.~Sui, V.~Zhuang, J.~Burdick, Y.~Yue, Stagewise safe {B}ayesian optimization
  with {G}aussian processes, in: International Conference on Machine Learning,
  2018, pp. 4781--4789.

\bibitem{wabersich2021predictive}
K.~P. Wabersich, M.~N. Zeilinger, A predictive safety filter for learning-based
  control of constrained nonlinear dynamical systems, Automatica 129 (2021)
  109597.

\bibitem{wieland2007constructive}
P.~Wieland, F.~Allg{\"o}wer, Constructive safety using control barrier
  functions, IFAC Proceedings Volumes 40~(12) (2007) 462--467.

\bibitem{cheng2019end}
R.~Cheng, G.~Orosz, R.~M. Murray, J.~W. Burdick, End-to-end safe reinforcement
  learning through barrier functions for safety-critical continuous control
  tasks, in: AAAI Conference on Artificial Intelligence, 2019, pp. 3387--3395.

\bibitem{SchoelkopfKernels}
B.~Sch\"{o}lkopf, A.~J. Smola, Learning with Kernels: Support Vector Machines,
  Regularization, Optimization, and Beyond, MIT Press, Cambridge, MA, USA,
  2001.

\bibitem{tight_bounds}
C.~Fiedler, C.~W. Scherer, S.~Trimpe, Practical and rigorous uncertainty bounds
  for {G}aussian process regression, AAAI Conference on Artificial Intelligence
  35~(8) (2021) 7439--7447.

\bibitem{puterman2014markov}
M.~L. Puterman, Markov decision processes: discrete stochastic dynamic
  programming, John Wiley \& Sons, 2014.

\bibitem{Rassmussen}
C.~E. Rasmussen, C.~K.~I. Williams, Gaussian Processes for Machine Learning
  (Adaptive Computation and Machine Learning), The MIT Press, 2005.

\bibitem{srinivas}
N.~Srinivas, A.~Krause, S.~M. Kakade, M.~W. Seeger, Information-theoretic
  regret bounds for {G}aussian process optimization in the bandit setting, IEEE
  Transactions on Information Theory 58~(5) (2012) 3250--3265.

\bibitem{pmlr-v70-chowdhury17a}
S.~R. Chowdhury, A.~Gopalan, On kernelized multi-armed bandits, in:
  International Conference on Machine Learning, 2017, pp. 844--853.

\bibitem{elementsofIT}
T.~M. Cover, J.~A. Thomas, Elements of Information Theory (Wiley Series in
  Telecommunications and Signal Processing), Wiley-Interscience, USA, 2006.

\bibitem{ContextualGPBO}
A.~Krause, C.~Ong, Contextual gaussian process bandit optimization, in:
  Advances in Neural Information Processing Systems, 2011.

\bibitem{DUIVENVOORDEN201711800}
R.~R. Duivenvoorden, F.~Berkenkamp, N.~Carion, A.~Krause, A.~P. Schoellig,
  Constrained {B}ayesian optimization with particle swarms for safe adaptive
  controller tuning, IFAC-PapersOnLine 50~(1) (2017) 11800--11807, 20th IFAC
  World Congress.

\bibitem{robotics_handbook}
B.~Siciliano, O.~Khatib, Springer Handbook of Robotics, 2nd Edition, Springer
  Publishing Company, Incorporated, 2016.

\bibitem{mujoco}
E.~Todorov, T.~Erez, Y.~Tassa, Mujoco: A physics engine for model-based
  control, in: IEEE/RSJ International Conference on Intelligent Robots and
  Systems, 2012, pp. 5026--5033.

\bibitem{bertsekas}
D.~P. Bertsekas, Dynamic Programming and Optimal Control, 2nd Edition, Athena
  Scientific, 2000.

\bibitem{racecarSafeopt}
A.~Wischnewski, J.~Betz, B.~Lohmann, A model-free algorithm to safely approach
  the handling limit of an autonomous racecar, in: IEEE International
  Conference on Connected Vehicles and Expo, 2019, pp. 1--6.

\bibitem{siemensSafeOpt}
M.~Fiducioso, S.~Curi, B.~Schumacher, M.~Gwerder, A.~Krause, Safe contextual
  {B}ayesian optimization for sustainable room temperature {PID} control
  tuning, in: International Joint Conference on Artificial Intelligence, 2019,
  pp. 5850--5856.

\bibitem{GooseApplied}
C.~König, M.~Turchetta, J.~Lygeros, A.~Rupenyan, A.~Krause, Safe and efficient
  model-free adaptive control via bayesian optimization, in: IEEE International
  Conference on Robotics and Automation, 2021, pp. 9782--9788.

\bibitem{CooperSafeOpt}
S.~E. Cooper, T.~I. Netoff, Multidimensional bayesian estimation for deep brain
  stimulation using the safeopt algorithm, medRxiv (2022).

\bibitem{rothfuss2022meta}
J.~Rothfuss, C.~Koenig, A.~Rupenyan, A.~Krause, Meta-learning priors for safe
  bayesian optimization, arXiv preprint arXiv:2210.00762 (2022).

\bibitem{hyperParamFelix}
F.~Berkenkamp, A.~P. Schoellig, A.~Krause, No-regret bayesian optimization with
  unknown hyperparameters, Journal of Machine Learning Research (2019) 1--24.

\bibitem{leggedrobotics}
A.~Schperberg, S.~D. Cairano, M.~Menner, Auto-tuning of controller and online
  trajectory planner for legged robots, IEEE Robotics and Automation Letters
  (2022).

\end{thebibliography}
\newpage
\appendix
\section{Proofs of Theoretical Results}\label{theory_results}

In this section, we provide proof for the theoretical results stated in the main body of the paper. In the following, we denote by $k$ discrete time indices and with $t$ continuous ones. This difference is important because, while we obtain state measurements at discrete times, we need to preserve safety at all times. Moreover, similarly to the notation in \gosafe~\cite{DBLP:gosafe}, we denote by $\xi_{(t,x(t),a)}= \{x(t) + \int_t^{t'} z(x(\tau);\pi^a(x(\tau)))\mathrm{d}\tau \mid t'\geq t\}$ all the states in the trajectory induced by the policy $a$ starting from $x(t)$ at time $t$.

\subsection{Safety Guarantees}\label{sec:BC_proof}

In the following, we prove \cref{thm:safety}, which gives the safety guarantees for \gosafeopt. Since \gosafeopt has two stages, \gls{LSE} and \gls{GE}, we can study their safety separately. For \gls{LSE}, \cite{DBLP:journals/corr/BerkenkampKS16} provides safety guarantees. Therefore, here we focus on the safety guarantees for \gls{GE} and then show that combining both will guarantee the safety of the overall algorithm.
To this end, we first make a hypothesis on our safe set $S_n$ and confidence bounds $l_n(a,i)$ and $u_n(a,i)$. 
\begin{hypothesis}
Let $S_n \neq \emptyset$. The following properties hold for all $i \in \mathcal{I}_g$, $n\geq 0$ with probability at least $1-\delta$:
\begin{align}
    &\forall a \in S_n: g_i(a,x_0) \geq 0, \\
    &\forall a \in \mathcal{A}: l_n(a,i) \leq g_i(a,x_0) \leq u_n(a,i).
\end{align}
\label{hyp:safeset}
\end{hypothesis}
\vspace{-2em}
We leverage this hypothesis to prove that we are safe during \gls{GE} and then we show that it is satisfied for \gosafeopt. Particularly, 
during \gls{LSE}, \cite{DBLP:journals/corr/BerkenkampKS16} proves that our hypothesis is fulfilled. Hence, before \gls{GE}, the safe set and the confidence intervals satisfy it. In the following, we show the updates of the safe sets and the confidence intervals implemented by  \gls{GE} also satisfy our hypothesis, which is sufficient to conclude that the hypothesis is satisfied  for all $n\geq 0$ (we will make this concrete in \cref{lemm:assumption_safeset_proof}).

\looseness=-1
During \gls{GE}, we receive measurements of the state in discrete times and evaluate our boundary condition to trigger a backup policy if necessary. Therefore, we first show that even with discrete-time measurements, we can still guarantee safety in continuous time.
\begin{lemma}
Let Assumptions~\ref{ass:one_step_jump} and~\ref{ass:constraint} hold and let $k_{+}\geq k_{-} \geq 0$ be arbitrary integers. If, for all integers $k \in [k_{-},  k_{+}]$, there exists $a_s \in \mathcal{A}$ such that $g_i(a_s,x(k)) \geq L_{\mathrm{x}} \Xi$ for all $i \in \mathcal{I}_g$, then $\bar{g}_i(x(t)) \geq 0$,  for all $t \in [k_{-}\Delta t, (k_{+}+1)\Delta t]$ and $i \in \mathcal{I}_g$.
\label{lemm:bar_g_safe}
\end{lemma}
\begin{proof}
 By choice of the sampling scheme, we have that the state $x(k)$ measured in discrete time, corresponds to the state $x(k\Delta t)$ in continuous time. Hence,
 $g_i(a_s,x(k))=g_i(a_s,x(k\Delta t))$.
 Consider some $k\geq 0$ and $a_s \in \mathcal{A}$ such that $g_i(a_s,x(k\Delta t)) \geq L_{\mathrm{x}} \Xi$.
 For any $t \in [k\Delta t, (k+1)\Delta t]$ we have
\begin{align*}
   g_i(a_s,x(k\Delta t)) -g_i(a_s,x(t)) &\leq L_{\mathrm{x}} \normAny{ x(k\Delta t) - x(t)} \tag{Lipschitz continuity (Assumption~\ref{ass:smoothness_assumption})}\\
   &\leq  L_{\mathrm{x}} \Xi. \tag{Assumption~\ref{ass:one_step_jump}}
\end{align*}
Now, since $g_i(a_s,x(k\Delta t)) \geq L_{\mathrm{x}} \Xi$, we have, for all $t \in [ k\Delta t, (k+1)\Delta t]$ and $i \in \mathcal{I}_g$,
\begin{equation}
    g_i(a_s,x(t)) \geq g_i(a_s,x(k\Delta t))- L_{\mathrm{x}} \Xi \geq 0. \label{eq:single_step_safety_condition}
\end{equation}
For our choice of constraints (Assumption~\ref{ass:constraint}) this implies $\bar{g}_i(x(t)) \geq 0$ for all $i\in \mathcal{I}_g$ and $t \in [ k\Delta t, (k+1)\Delta t]$. 
Finally, since this holds for all integers $k$ with $k_{-} \leq k \leq k_{+}$, it also holds for all $t \in [k_{-}\Delta t, (k_{+}+1)\Delta t]$.
\end{proof}
Now we have established a condition that guarantees for a given time interval that $\bar{g}_i(x(t)) \geq 0$ for all $i \in \mathcal{I}_g$. 

We collect parameter and state combinations during rollouts in our set of backups $\mathcal{B}_n$. The intuition here is that for a Markovian system, all states visited during a safe experiment are also safe. This is important as it allows \gosafeopt to learn backup policies for multiple states without actively exploring the state space. We formalize this in the following proposition.
\begin{proposition}
\label{prop:backup_policies}
Let Assumption~\ref{ass:constraint} hold. If $(a,x_0)$ is safe, that is, 
$\min_{x' \in \xi_{(0,x_0,a)}}\bar{g}_i(x')$ $\geq 0$ for all $i \in \mathcal{I}_g$, then, for all $t_1 \geq 0$, $\left(a,x(t_1)\right)$ is also safe, that is $\min_{x' \in \xi_{(t_1,x(t_1),a)}}\bar{g}_i(x') \geq 0$ for all $i \in \mathcal{I}_g$.
\end{proposition}

\begin{proof}
 The system in \cref{eq:syseq} is Markovian, i.e., for any $x(t_1)\in \xi_{(0,x_0,a)}$ and $x(t_2)\in \xi_{(0,x_0,a)}$ with $t_2>t_1>0$, 
 \begin{align*}
        x(t_2) 
        &= x_0 + \int\limits_{0}^{t_1}z(x(t);\pi^a(x(t))\mathrm{d}t+\int\limits_{t_1}^{t_2}z(x(t);\pi^a(x(t))\mathrm{d}t \\
        &= x(t_1) + \int\limits_{t_1}^{t_2}z(x(t);\pi^a(x(t))\mathrm{d}t.
    \end{align*}
Therefore, a trajectory starting in $x(t_1)$ will always result in the same state evolution, independent of how we arrived at $x(t_1)$.
Combining this and Assumption~\ref{ass:constraint}, we get
\begin{align*}
    g_i(a,x(t_1)) &= \min_{x' \in \xi_{(t_1,x(t_1),a)}}\bar{g}_i(x')\tag*{Assumption~\ref{ass:constraint}}\\
    &\geq \min_{x' \in \xi_{(0,x_0,a)}}\bar{g}_i(x')\ \tag*{Markov Property}\\
    &= g_i\left(a,x_0\right)\tag*{Assumption~\ref{ass:constraint}}\\
    &\geq 0.
\end{align*}
\end{proof}
In the following, we show that $g_i(a_s,x_0)$ is a lower bound for all points $(a_s,x_s)$ in $\mathcal{B}_n$, i.e., $g_i(a_s,x_s) \geq g_i(a_s,x_0)$. This will play a crucial role in showing that we preserve safety whenever we trigger a backup policy.
\begin{corollary}
Let Assumption~\ref{ass:constraint} hold. For all points $(a_s,x_s)$ in $\mathcal{B}_n$, $g_i(a_s,x_s) \geq g_i(a_s,x_0)$ for all $i \in \mathcal{I}_g$. 
\label{cor:backup_bound}
\end{corollary}
\begin{proof}
Each point $(a_s,x_s)$ in $\mathcal{B}_n$ is collected during a safe experiments (see \cref{alg:LSE} line~\ref{lst:line:update_GP_lse} and \cref{alg:GE} line~\ref{lst:line:update_GP_ge}). Therefore, $x_s \in \xi_{(0,x_0,a_s)}$. The result then follows from \cref{prop:backup_policies}.
\end{proof}
\cref{cor:backup_bound} shows that $l_n(a_s,i)$ is a conservative lower bound on $g_i(a_s,x_s)$. Crucially, if we can observe not just the rollouts but also the constraint values $g_i(a_s,x_s)$, we could model them with a \gls{GP} to obtain a potentially less conservative lower bound. However, in our work, we only assume that we can measure $g_i(a_s,x_0)$ (Assumption \ref{ass:observation_model}). 

\cref{prop:backup_policies} and \cref{cor:backup_bound} formalize how we collect our backup policies and leverage them in our boundary condition. 
In the following, we prove that experiments, where we trigger a backup policy, are safe. First, we show that
if the boundary condition is triggered at a time step $k^{*}$, then we are safe up until $ k^{*} \Delta t$, i.e., time of trigger.
\begin{lemma}
Let the assumptions from \cref{thm:safety} and Hypothesis~\ref{hyp:safeset} hold. If, during \gls{GE}, the boundary condition from \cref{BC_algo} triggers a backup policy at time step $k^{*} > 0$, then, for all $t \leq k^{*} \Delta t $ and $i \in \mathcal{I}_g$, $\Bar{g}_i(x(t)) \geq 0$ with probability at least $1-\delta$.
\label{lemm:safe_tolerance}
\end{lemma}
\begin{proof}
Consider $k < k^*$. Since the boundary condition (\cref{BC_algo}) did not trigger a backup policy at $k$, we have
\begin{equation}
\label{eqn:safe_backup_prop}
    \exists (a_s,x_s) \in \mathcal{B}_n \text{ such that } l_n(a_s,i)  \geq L_{\mathrm{x}} \left(\normAny{x(k) - x_s}+\Xi\right), \forall i \in  \mathcal{I}_g.
\end{equation}
By Lipschitz continuity of $g$, we have
\begin{equation}
    g_i(a_s,x_s) - g_i(a_s,x(k)) \leq L_{\mathrm{x}}\normAny{x(k) - x_s},
\end{equation}
which implies
\begin{align*}
    g_i(a_s,x(k)) &\geq g_i(a_s,x_s) - L_{\mathrm{x}}\normAny{x(k) - x_s}\\
    &\geq g_i(a_s,x_0) - L_{\mathrm{x}}\normAny{x(k) - x_s}\tag*{\cref{cor:backup_bound}}\\
    &\geq l_n(a_s,i) - L_{\mathrm{x}}\normAny{x(k) - x_s} \tag*{Hypothesis~\ref{hyp:safeset}}\\
     &\geq L_{\mathrm{x}}\Xi\tag*{\eqref{eqn:safe_backup_prop}}
\end{align*}
for all $i\in\mathcal{I}_g$ and $k < k^{*}$.
Therefore,  we can use \cref{lemm:bar_g_safe} to prove the claim by choosing $k_{-}=0$ and $k_{+} = k^{*} - 1$. 

\end{proof}

\cref{lemm:safe_tolerance} shows that up until the time we trigger our boundary condition, we are safe with enough tolerance ($L_{\mathrm{x}}\Xi$) to guarantee safety. 
In the following, we show that if we trigger a safe backup policy at $k^{*}$, we will fulfill our constraints for all times after triggering. 

\begin{lemma}
Let the assumptions from \cref{thm:safety} and Hypothesis~\ref{hyp:safeset} hold. If during a \gls{GE} experiment with parameter $a_{\textrm{\gls{GE}}}$ at time step $k^{*} \geq 0$, our boundary condition triggers the backup policy $a^{*}_s$ defined as  (see \cref{eq:backup_action})
\begin{equation}
    a_s^* = \underset{\left\{a \in \mathcal{A} | \exists x \in \mathcal{X}; (a,x) \in \mathcal{B}_n\right\}}{\arg\max} \; \underset{i \in \mathcal{I}_g}{\min} \; l_n(a,i) - L_\mathrm{x}\normAny{x-x(k^{*})}.
    \label{eq:backup_policy_formula}
\end{equation}
Then for all $t \geq  k^{*}\Delta t$, $\Bar{g}_i(x(t)) \geq 0$ for all $i \in \mathcal{I}_g$ with probability at least $1-\delta$.
\label{lemm:safe_trigger_after}
\end{lemma}
\begin{proof}
We want to show that \cref{eq:backup_policy_formula} finds a parameter $a^*_s$ such that $g_i(a^*_s, x(k^{*})) \geq 0$. For $k^* = 0$, this follows by definition because $\mathcal{B}_n$ consists of safe rollouts (see \cref{alg:LSE} line~\ref{lst:line:update_GP_lse} and \cref{alg:GE} line~\ref{lst:line:update_GP_ge}) and thus, for all parameters $a_s$ in $\mathcal{B}_n$ and $i \in \mathcal{I}_g$, we have $g_i(a_s,x_0) \geq 0$.

Let us now consider any integer $k^* > 0$. Let $(a_s, x_s) \in \mathcal{B}_n$ be arbitrary.
Following the same Lipschitz continuity-based arguments as in ~\cref{lemm:safe_tolerance},
we have for all $i\in\mathcal{I}_g$, :
\begin{align*}
    g_i(a_s,x(k^{*})) & \geq
    l_n(a_s,i) - L_\mathrm{x}\normAny{x_s-x(k^{*})} \tag*{same as~\cref{lemm:safe_tolerance}} \\
    &\geq l_n(a_s,i) - L_{\mathrm{x}} \big(\normAny{x(k^{*}-1) - x_s}+  \normAny{x(k^{*})-x(k^{*}-1)}\big) \tag{Triangle inequality}\\
    %
    &\geq l_n(a_s,i) - L_{\mathrm{x}} \left(\normAny{x(k^{*}-1) - x_s}+ \Xi\right)
    \tag{Assumption~\ref{ass:one_step_jump}} \\
    &\geq 0, \numberthis \label{lower_bound_pos}
\end{align*}
where the last inequality follows from the fact that the boundary condition was not triggered at time step $k^{*} - 1$ (see \cref{sec:global_exploration}).
Furthermore, from~\cref{lower_bound_pos} we can conclude that there exists $a_s \in \mathcal{A}$ such that for some $x_s \in \mathcal{X}$, $(a_s,x_s) \in \mathcal{B}_n$, and $l_n(a_s,i) - L_\mathrm{x}\normAny{x_s-x(k^{*})} \geq 0$ for all $i\in \mathcal{I}_g$. Therefore, we have for $a^*_s$ recommended by \cref{eq:backup_policy_formula}:
\begin{equation*}
\underset{\left\{a \in \mathcal{A} | \exists x \in \mathcal{X}; (a,x) \in \mathcal{B}_n\right\}}{\max} \; \underset{i \in \mathcal{I}_g}{\min} \; l_n(a,i) - L_\mathrm{x}\normAny{x-x(k^{*})} \geq 0.
\end{equation*}
Hence, $g_i(a_s^*, x(k^{*})) \geq 0$ for all $i \in \mathcal{I}_g$ with probability at least $1-\delta$, which proves the claim.
\end{proof}

\cref{lemm:safe_tolerance,lemm:safe_trigger_after} show that, if we trigger a backup policy during \gls{GE}, we can guarantee the  safety of the experiment before and after switching to the backup policy, respectively.

Next, we prove that, if the backup policy is not triggered during \gls{GE} with parameter $a_{\textrm{\gls{GE}}}$, then $a_{\textrm{\gls{GE}}}$  is safe with high probability.

\begin{lemma}
Let the assumptions from \cref{thm:safety} and Hypothesis~\ref{hyp:safeset} hold. If, during \gls{GE} with parameter $a_{\textrm{\gls{GE}}}$, a backup policy is not triggered by our boundary condition, then $a_{\textrm{\gls{GE}}}$  is safe with probability at least $1-\delta$, that is, $g_i(a_{\textrm{\gls{GE}}},x_0) \geq 0$ for all $i \in \mathcal{I}_g$.
\label{lemm:safe_param}
\end{lemma}
\begin{proof}

Assume the experiment was not safe, i.e., there exists a $t \geq 0$, such that for some $i \in \mathcal{I}_g$ $\bar{g}_i(x(t)) < 0$. Consider the time step $k\geq 0$ such that $t \in [k\Delta t, (k+1)\Delta t ]$. Since the boundary condition was not triggered during the whole experiment, it was also not triggered at time step $k$. This implies that (see \cref{sec:global_exploration}) there exists a point $(a_s,x_s) \in \mathcal{B}_n$ such that
\begin{equation}
    l_n(a_s,i) - L_{\mathrm{x}}\left(\normAny{x_s-x(k)}+\Xi\right) \geq 0,
\end{equation}
for all $i \in \mathcal{I}_g$. Therefore, we have $g_i(a_s,x(k)) \geq L_x \Xi$ (Hypothesis~\ref{hyp:safeset}). Hence, from \cref{lemm:bar_g_safe} we have $\bar{g}_i(x(t)) \geq 0$ for all $i \in \mathcal{I}_g$.
This contradicts our assumption that for some $t\geq 0$ and $i \in \mathcal{I}_g$,  $\bar{g}_i(x(t)) < 0$.
\end{proof}

The following Corollary summarizes the safety of \gls{GE}.

\begin{corollary}
\label{cor:safe_GE}
Under the assumptions from \cref{thm:safety} and Hypothesis~\ref{hyp:safeset} \gosafeopt is safe during \gls{GE}, i.e., for all $t \geq 0$, $\Bar{g}_i(x(t)) \geq 0$ for all $i\in \mathcal{I}_g$. 
\end{corollary}
\begin{proof}
Two scenarios can occur during \gls{GE}, 
 (\emph{i}) a backup policy is triggered at some time step $k^{*} \geq 0$, (\emph{ii}) the experiment is completed without triggering a backup policy.
For the first case, Lemma~\ref{lemm:safe_trigger_after} guarantees that we are safe after triggering the backup policy, and \cref{lemm:safe_tolerance} guarantees that we are safe before we trigger the backup. For second scenario, \cref{lemm:safe_param} guarantees safety. 
\end{proof}
We have now shown that under the assumptions of \cref{thm:safety} combined with Hypothesis~\ref{hyp:safeset}, we can guarantee that we are safe during \gls{GE}, irrespective of whether we trigger a backup policy or not. 
We leverage this result to show that Hypothesis~\ref{hyp:safeset} is satisfied for \gosafeopt.

\begin{lemma}
Let the assumptions from \cref{thm:safety} hold and $\beta_n$ be defined as in \cite{DBLP:journals/corr/BerkenkampKS16}. Then, Hypothesis~\ref{hyp:safeset} is satisfied for \gosafeopt, that is, with probability at least $1-\delta$ for all $i \in \mathcal{I}_g$ and $n\geq 0$
\begin{align}
    &\forall a \in S_n: g_i(a,x_0) \geq 0, \label{eq:ass_safeset} \\
    &\forall a \in \mathcal{A}: l_n(a,i) \leq g_i(a,x_0) \leq u_n(a,i).
    \label{eq:ass_bounds}
\end{align}
\label{lemm:assumption_safeset_proof}
\end{lemma}
\begin{proof}
We use induction on $n$. 
\begin{itemize}[leftmargin=*]
    \item[] \emph{Base case $n=0$:} By Assumption~\ref{ass:safe_seed},  we have, for all $a \in S_0$, $g_i(a,x_0) \geq 0$ for all $i \in \mathcal{I}_g$. Moreover, the initialization of the confidence intervals presented in \cref{safeopt_gosafe} is as follows: $l_0(a,i)=0$ if $a \in S_0$ and $-\infty$ otherwise, and $u_0(a,i)=\infty$ for all $a\in \mathcal{A}$. Thus, it follows that $l_0(a,i) \leq g_i(a,x_0) \leq u_0(a,i)$ for all $a \in \mathcal{A}$.
    \item[] \emph{Inductive step:} Our induction hypothesis is $l_{n-1}(a,i) \leq g_i(a,x_0) \leq u_{n-1}(a,i)$ and $g_i(a,x_0) \geq 0$ for all $a\in S_{n-1}$ and for all $i \in \mathcal{I}_g$. Based on this, we prove that these relations hold for iteration $n$.
    
    We start by showing that $l_n(a,i) \leq g_i(a,x_0) \leq u_n(a,i)$ for all $a \in \mathcal{A}$. To this end, we distinguish between the different updates of the two stages of \gosafeopt, \gls{LSE} and \gls{GE}. During \gls{LSE}, we define $l_n(a,i)$ and  $u_n(a,i)$ as
    \begin{align*}
        l_n(a,i) &=  \max({l_{n-1}(a,i), \mu_{n}(a,i) - \beta_{n}\sigma_{n}(a,i)}), \\
        u_n(a,i) &=  \min({u_{n-1}(a,i), \mu_{n}(a,i) + \beta_{n}\sigma_{n}(a,i)}).
    \end{align*}
    We know that $g_i(a,x_0) \geq l_{n-1}$ by induction hypothesis and $g_i(a,x_0) \geq \mu_{n}(a,i) - \beta_{n}\sigma_{n}(a,i)$ with probability $1-\delta$ from \cite{DBLP:journals/corr/BerkenkampKS16}. This implies $g_i(a,x_0) \geq l_{n}$. A similar argument holds for the upper bound.
    
    During \gls{GE}, we update $l_n(a,i)$ if the parameter we evaluate induces a trajectory that does not trigger a backup policy (see \cref{alg:GE} line~\ref{lst:line:update_S_ge}). For this parameter, the induction hypothesis allows us to use  \cref{lemm:safe_param} and conclude $g_i(a,x_0) \geq 0$. Therefore, the update of the confidence intervals during \gls{GE} also satisfies \cref{eq:ass_bounds} for iteration $n$, thus completing the induction step for the confidence intervals.
    
    As for the confidence intervals, we distinguish between the different updates of the safe set implemented by \gls{LSE} and \gls{GE}. In the case of \gls{GE}, we update the safe set by adding the evaluated policy parameter $a$ only if it does not trigger a backup, i.e., $S_n = S_{n-1} \cup \{a\}$. Following the same argument as above, we can conclude $g_i(a,x_0)\geq 0$ for all $i \in \mathcal{I}_g$. This together with the induction hypothesis means $g_i(a,x_0) \geq 0$ for all $i \in \mathcal{I}_g$ and $a \in S_n$ in case of a \gls{GE} update.
    
    Now we focus on \gls{LSE}.  We showed \cref{eq:ass_bounds} holds for $n$. Moreover, we know by induction hypothesis $g_i(a,x_0) \geq 0$ for all $a \in S_{n-1}$ and for all $i \in \mathcal{I}_g$ with high probability. The update equation for the safe set (\cref{eq:safeset}) gives for all $a' \in S_n \setminus S_{n-1}$, there exists $a \in S_{n-1}$ such that for all $i \in \mathcal{I}_g$
    \begin{equation}
        l_n(a,i) - L_{\mathrm{a}}\normAny{a-a'} \geq 0. \label{eq:safe_set_redef}
    \end{equation}
    We show that this is enough to guarantee with high probability that $g_i(a',x_0) \geq 0$. 
    Due to the Lipschitz continuity of the constraint functions, we have 
    \begin{align*}
        g_i(a',x_0) &\geq g_i(a,x_0) - L_{\mathrm{a}}\normAny{a-a'}, \\
        &\geq l_n(a,i) - L_{\mathrm{a}}\normAny{a-a'} \geq 0 \tag{\cref{eq:safe_set_redef}}.
    \end{align*}
    Therefore, $g_i(a',x_0) \geq 0$ for all $i \in \mathcal{I}_g$ and $a \in S_n$ with probability at least $1-\delta$ also in case of an \gls{LSE} step.
\end{itemize}
\end{proof}
 Lemma~\ref{lemm:assumption_safeset_proof} ensures that Hypothesis~\ref{hyp:safeset} holds for \gosafeopt. We also know that under the same assumption as \cref{thm:safety} and Hypothesis~\ref{hyp:safeset}, we are safe during \gls{GE} (see \cref{cor:safe_GE}). Hence, we can now guarantee safety during \gls{GE}.
Finally, we prove \cref{thm:safety}, which guarantees safety for \gosafeopt. 
\safegosafeopt*
\begin{proof}
We perform \gosafeopt in two stages; \gls{LSE} and \gls{GE}. In Lemma~\ref{lemm:assumption_safeset_proof}, we proved that for all parameters $a \in S_n$, $g_i(a,x_0) \geq 0$ for all $i \in \mathcal{I}_g$ with probability at least $1 -\delta$. During \gls{LSE} we query parameters from $S_n$ (\cref{eq:acq_lse}). Therefore, the experiments are safe. During \gls{GE}, \cref{cor:safe_GE} proves that when assumptions from \cref{thm:safety} and Hypothesis~\ref{hyp:safeset} hold, we are safe during \gls{GE} for our choice of $\beta_n$. Furthermore, in Lemma~\ref{lemm:assumption_safeset_proof} we proved that Hypothesis~\ref{hyp:safeset} is satisfied for \gosafeopt. Hence, we can conclude that if the assumptions from \cref{thm:safety} hold, we are safe during \gls{GE} at all times.
\end{proof}
\subsubsection{Proof of Boundary Condition For Noisy Measurements}\label{noisy_case_bc}
\begin{lemma}
Assume at each time step $k$ we receive a noisy measurement of the state to evaluate our boundary condition, i.e., $y = x + \varepsilon$ and $\varepsilon$ \gls{iid} Specifically, assume $P(\normAny{\varepsilon} \leq \sfrac{d}{2}) \geq \sqrt{1-\delta_2}$. If we have for some $(a_s, y_s) \in \mathcal{B}_n$ ($y_s = x_s + \varepsilon_s$)
\begin{equation*}
   l_n(a_s, i) - L_{\mathrm{x}}(\normAny{y - y_s} + \Xi + d) \geq 0,
\end{equation*}
then
with probability at least $1-\delta_2$ we have
\begin{equation*}
   l_n(a_s, i) - L_{\mathrm{x}}(\normAny{x - x_s} + \Xi) \geq 0.
\end{equation*}
\end{lemma}
\begin{proof}
We would like to show that 
\begin{equation*}
   l_n(a_s, i) - L_{\mathrm{x}}(\normAny{y - y_s} + \Xi + d)  \leq l_n(a_s, i) - L_{\mathrm{x}}(\normAny{x - x_s} + \Xi).
\end{equation*}
This implies that 
$d \geq \normAny{x-x_s} - \normAny{y-y_s}$. 

Accordingly, 
\begin{align*}
    \normAny{x-x_s} - \normAny{y-y_s} &\leq  \normAny{x-x_s - (y-y_s)} \tag{reverse triangle inequality}\\
    &= \normAny{\varepsilon_s - \varepsilon} \leq \normAny{\varepsilon} + \normAny{\varepsilon_s} \\
    &\leq d \tag{with probability at least $1-\delta_2$.}
\end{align*}
\end{proof}
Following the lemma, we can come up with a more conservative boundary condition (with one step jump bound $\Xi' = \Xi + d)$ which still guarantees safety. However, the price we pay for not measuring our state perfectly is the additional probability term $1-\delta_2$. Lastly, here we only look at the influence of noisy state measurements on the boundary condition. Nevertheless, if the policy $\pi$ uses some form of feedback, the noise also enters the dynamics. In this work, we assume that this influence is captured by our observation model, see Assumption~\ref{ass:observation_model}.
\subsection{Optimality Guarantees}\label{sec:proof_discoverable}
In this section, we prove \cref{thm:optimality} which guarantees that the safe global optimum can be found with $\epsilon$-precision if it is discoverable at some iteration $n\geq 0$ (see \cref{def:discoverable_set}). 
Then, we show in  \cref{lemm:discoverable} that for many practical applications, this discoverability condition is satisfied.
\subsubsection[Proof of Optimality Theorem]{Proof of \cref{thm:optimality}}
We first define the largest region that \gls{LSE} can safely explore for a given safe initialization $S$ and then we show that we can find the optimum with $\epsilon$-precision within this region.
To this end, we define the reachability operator $R^c_{\epsilon}(S)$  and the fully connected safe region $\bar{R}^c_{\epsilon}(S)$ by (adapted from~\cite{DBLP:journals/corr/BerkenkampKS16, DBLP:gosafe})
\begin{align}
    R^c_{\epsilon}(S) \coloneqq S \cup \{ a \in \mathcal{A} \mid \exists a' \in S \text{ such that } &g_i(a',x_0) - \epsilon - L_a \normAny{a-a'} \geq 0, \notag \\ 
    &\forall i \in \mathcal{I}_g\}, \label{eq:reachability_operator}
    \end{align} 
    \vspace{-2em}
    \begin{equation}
    \Bar{R}^c_{\epsilon}(S) \coloneqq  \lim_{n \to \infty} \left(R^c_{\epsilon}\right)^n(S). \label{eq:connected_set}
    \end{equation}
The reachability operator $R^c_{\epsilon}(S)$ contains the parameters we can safely explore if we know our constraint function with $\epsilon$-precision within some safe set of parameters $S$. Further, $(R^c_\epsilon)^n(S)$ denotes the repeated composition of $R^c_{\epsilon}(S)$ with itself, and $\Bar{R}^c_{\epsilon}(S)$ its closure. Next, we derive a property for the reachability operator, that we will leverage to provide optimality guarantees.
\begin{lemma}\label{lemm:properties_Rcon}
Let $A\subseteq S$, if $\Bar{R}^{c}_{\epsilon}(A)\setminus S \neq \emptyset$, then $R^{c}_{\epsilon}(S)\setminus S \neq \emptyset$.
\end{lemma}
\begin{proof}
    This lemma is a straightforward generalization of \cite[Lem.~7.4]{DBLP:journals/corr/BerkenkampKS16}. Assume $R^c_{\epsilon}(S) \setminus S = \emptyset$, we want to show that this implies $\Bar{R}^{c}_{\epsilon}(A)\setminus S = \emptyset$. By definition $R^c_{\epsilon}(S) \supseteq S$ and therefore $R^c_{\epsilon}(S) = S$. Iteratively applying $R^c_{\epsilon}$ to both the sides, we get in the limit $\bar{R}^c_{\epsilon}(S) = S$.
    Furthermore, because $A \subseteq S$, we have $\bar{R}^{c}_{\epsilon}(A) \subseteq \Bar{R}^c_{\epsilon}(S)$~\cite[Lem.~7.1]{DBLP:journals/corr/BerkenkampKS16}. Thus, we obtain $\bar{R}^{c}_{\epsilon}(A) \subseteq \Bar{R}^c_{\epsilon}(S)=S$, which leads to $\Bar{R}^{c}_{\epsilon}(A)\setminus S = \emptyset$. 
\end{proof}
In the following, we prove that our \gls{LSE} convergence criterion (see \cref{eq:conv_lse}) guarantees that for the safe initialization $S$, we can explore $\Bar{R}^c_{\epsilon}(S)$ during \gls{LSE} in finite time.
\begin{theorem}
\label{prop:LSE_optimality}
Consider any  $\epsilon>0$ and $\delta>0$. Let Assumptions~\ref{ass:smoothness_assumption} and~\ref{ass:observation_model} hold, $\beta_n$ be defined as in \cite{DBLP:journals/corr/BerkenkampKS16}, and $S\subseteq \mathcal{A}$ be an initial safe seed of parameters, i.e., $g(a,x_0)\geq 0$ for all $a \in S$. Assume that the information gain $\gamma_n$ grows sublinearly with $n$ for the kernel $k$. Further let $n^*$ be the smallest integer such that (cdf.\ the convergence criterion of \gls{LSE} in~\cref{eq:conv_lse})
\begin{equation}
    \underset{a \in \mathcal{G}_{n^*-1} \cup \mathcal{M}_{n^*-1} }{\max} \underset{i \in \mathcal{I}}{\max } \; w_{n^*-1}(a,i) < \epsilon \text{ and } S_{n^*-1} = S_{n^*}. \label{stopping_criteria}
\end{equation}
Then we have that $n^*$ is finite and 
when running \gls{LSE}, the following holds with probability at least $1-\delta$ for all $n\geq n^*$:
\begin{align}
    &\Bar{R}^c_{\epsilon}(S) \subseteq S_{n}, \label{eq:conv_safeset} \\
  &f(\hat{a}_n) \geq \max_{a \in \Bar{R}^c_{\epsilon}(S)} f(a) -\epsilon, \label{eq:conv_opt}
\end{align}
with $\hat{a}_n = \underset{a \in S_n}{\arg\max} \; l_n(a,0)$. 
\end{theorem}
\begin{proof}
We first leverage the result from \cite[Thm.~4.1]{DBLP:journals/corr/BerkenkampKS16} which provides the following \emph{worst-case} bound on $n^*$ 
\begin{equation}
    \frac{n^*}{\beta_{n^*} \gamma_{|\mathcal{I}|n^*}} \geq \frac{C_1\left(\Bar{R}^c_{0}(S)\right)+1}{\epsilon^2}, \label{eq:recommended_optimum_lse}
\end{equation}
where $C_1 = 8/\log(1+\sigma^{-2})$ and $n^*$ is the smallest integer that satisfies \cref{eq:recommended_optimum_lse}. Hence, we have that $n^*$ is finite. The sublinear growth of $\gamma_n$ with $n$ is satisfied for many practical kernels, like the ones we consider in this work~\cite{srinivas}. 
Next, we prove \cref{eq:conv_safeset}. For the sake of contradiction, assume $\Bar{R}^c_{\epsilon}(S) \setminus S_{n^{*}} \neq \emptyset$. 
This implies, $R^c_{\epsilon}(S_{n^*}) \setminus S_{n^*} \neq \emptyset$ (\cref{lemm:properties_Rcon}). Therefore, there exists some $a \in \mathcal{A}\setminus S_{n^*}$ such that for some $a' \in S_{n^*}=S_{n^*-1}$ (\cref{stopping_criteria}), we have for all $i \in \mathcal{I}_g$
\begin{align*}
    0&\leq g_i(a',x_0) - \epsilon- L_{\mathrm{a}}\normAny{a-a'},\\
    &\leq u_{n^*-1}(a',i) - L_{\mathrm{a}}\normAny{a-a'} \tag*{(\cref{lemm:assumption_safeset_proof})}.
\end{align*}

Therefore, $a' \in \mathcal{G}_{n^*-1}$ (see \cite{DBLP:journals/corr/BerkenkampKS16} or \cref{sec:additional_defs} \cref{def:expanders}) and accordingly, $ w_{n^*-1}(a',i) < \epsilon $. Next, because $ w_{n^*-1}(a',i) < \epsilon $ ,we have for all $i \in \mathcal{I}_g$
\begin{equation} 
    0\leq g_i(a',x_0) - \epsilon- L_{\mathrm{a}}\normAny{a-a'} \leq l_{n^*-1}(a',i) - L_{\mathrm{a}}\normAny{a-a'}.
\end{equation}
This means $a \in S_{n^*}$ (\cref{eq:safeset}), which is a contradiction. Thus, we conclude  that $\Bar{R}^c_{\epsilon}(S) \subseteq S_{n^{*}}$ and because $S_{n^*} \subseteq S_n$ for all $n\geq n^*$ (\cref{nondecreasing_propertyS}), we get $\Bar{R}^c_{\epsilon}(S) \subseteq S_{n}$.

Now we prove \cref{eq:conv_opt}. Consider any $n \geq n^*$. Note,  $w_{n^*-1}(a',i) < \epsilon $, implies $w_{n}(a',i) < \epsilon$ (see \cref{alg:LSE} line~\ref{lst:line:update_S_lse} or \cref{alg:GE} line~\ref{lst:line:update_S_ge}). For simplicity, we denote the solution of $\argmax_{a \in \Bar{R}^c_{\epsilon}(S)} f(a)$ as $a^*_{S}$. We have
\begin{align*}
    u_{n}(a^*_S,0) &\geq f(a^*_S) \tag{\cref{lemm:assumption_safeset_proof}} \\
    &\geq f(\hat{a}_{n}) \tag{by definition of $a^*_S$} \\
    &\geq  l_n(\hat{a}_{n},0) \tag{\cref{lemm:assumption_safeset_proof}} \\
    &= \max_{a \in S_{n}} l_{n}(a,0) \tag{by definition of $\hat{a}_n$}. \\
\end{align*}
Therefore, $a^*_S$ is a maximizer, i.e., $a^*_S \in \mathcal{M}_{n}$ (see \cref{sec:additional_defs} \cref{def:maximizers}) and has uncertainty less than $\epsilon$, that is, $w_{n}(a^*_S,i) < \epsilon $. Now, we show that $f(\hat{a}_{n}) \geq f(a^*_S) - \epsilon$. For the sake of contradiction assume, 
\begin{equation}
    f(\hat{a}_{n}) < f(a^*_S) - \epsilon. \label{eq:wrong_assumption}
\end{equation}
Then we obtain,
\begin{align*}
    l_{n}(a^*_S,0) &\leq l_{n}(\hat{a}_{n},0) \tag{by definition of $\hat{a}_{n}$} \\
    &\leq f(\hat{a}_{n}) \tag{\cref{lemm:assumption_safeset_proof}} \\
    &< f(a^*_S) - \epsilon \tag{by \cref{eq:wrong_assumption}}\\
    &\leq u_{n}(a^*_S,0) - \epsilon \tag{\cref{lemm:assumption_safeset_proof}} \\
    &\leq u_{n}(a^*_S,0) - w_{n}(a^*,0) \tag{because $w_{n}(a^*_S,0) \leq \epsilon$} \\
    &= l_{n}(a^*_S,0) \tag{by definition of $w_{n}(a^*_S,0)$},
\end{align*}
which is a contradiction. Therefore, we have $ f(\hat{a}_{n}) \geq f(a^*_S) - \epsilon$.
\end{proof}
\cref{prop:LSE_optimality} states that for a given safe seed $S$, the convergence of \gls{LSE} (\cref{eq:conv_lse}) implies that we have discovered its fully connected safe region $\Bar{R}^c_{\epsilon}(S)$ and recovered the optimum within the region with $\epsilon$-precision. 

Based on the previous results, we can show that if the safe global optimum is discoverable for some iteration $n\geq 0$ (see \cref{def:discoverable_set}), 
then we can find an approximately optimal safe solution. 
However, to prove optimality, what we also require is that if $a^* \in S_n$ then $a^* \in S_{n+1}$. 
\begin{proposition}
Let the assumptions from~\cref{thm:safety} hold. For any $n \geq 0$, the following property is satisfied for $S_n$.
\begin{equation}
   S_{n} \subseteq S_{n+1}, \label{nondecreasing_safeset}
\end{equation}
\label{nondecreasing_propertyS}
\end{proposition}
\begin{proof}
The safe set provably increases during \gls{LSE}~\cite[Lem.~7.1]{DBLP:journals/corr/BerkenkampKS16}. During \gls{GE}, the safe set is only updated if a new safe parameter is found. The proposed update also has the non-decreasing property (see \cref{alg:GE}, line~\ref{lst:line:update_S_ge}). Hence, we can conclude that $S_{n} \subseteq S_{n+1}$. 
\end{proof}
\cref{nondecreasing_propertyS} shows that if the safe global optimum $a^* \in S_n$, then $a^* \in S_{n+1}$. Next, we prove that if a new safe region $A$ is added to our safe set $S_n$, we will explore its largest reachable safe set $\bar{R}^c_{\epsilon}(A)$. 
\begin{lemma}
Consider any integer $n\geq 0$. Let $S_n$ be the safe set of parameters explored after $n$ iterations of \gosafeopt and let $\beta_n$ be defined as in \cite{DBLP:journals/corr/BerkenkampKS16}.
Consider $A = S_{n+1}\setminus S_{n}$. If $A \neq \emptyset$, then there exists a finite integer $\bar{n}>n$ such that $\Bar{R}^{\epsilon}_{c} (A) \cup \Bar{R}^{\epsilon}_{c} (S_n) \subseteq S_{\bar{n}}$ with probability at least $1-\delta$.
\label{lemma:explore_safe_region_after_ge}
\end{lemma}
\begin{proof}
First,  if $\Bar{R}^{\epsilon}_{c} (A) \setminus S_{\bar{n}} = \emptyset$ and 
$\Bar{R}^{\epsilon}_{c} (S_n) \setminus S_{\bar{n}} = \emptyset$ then
$\Bar{R}^{\epsilon}_{c} (A) \cup \Bar{R}^{\epsilon}_{c} (S_n) \subseteq S_{\bar{n}}$. We now show that  $\Bar{R}^{\epsilon}_{c} (A) \setminus S_{\bar{n}} = \emptyset$.
Assume that $\Bar{R}^{\epsilon}_{c} (A) \setminus S_{\bar{n}} \neq \emptyset$. We know that $A \subseteq S_{n+1} \subseteq S_{\bar{n}}$ (\cref{nondecreasing_propertyS}). This implies $R^c_{\epsilon}(S_{\bar{n}}) \setminus S_{\bar{n}} \neq \emptyset$ (\cref{lemm:properties_Rcon}). 
Since $A \neq \emptyset$, the safe set is expanding. For \gosafeopt, this can either happen during \gls{LSE} or during \gls{GE} when a new parameter is successfully evaluated, i.e., the boundary condition is not triggered. In either case, we perform \gls{LSE} till convergence. 
Let $\bar{n} > n$ be the smallest integer for which we converge during \gls{LSE}, i.e., for which 
\begin{equation}
    \underset{a \in \mathcal{G}_{\bar{n}-1} \cup \mathcal{M}_{\bar{n}-1} }{\max} \underset{i \in \mathcal{I}}{\max } \; w_{\bar{n}-1}(a,i) < \epsilon \text{ and } S_{\bar{n}-1} = S_{\bar{n}}
\end{equation}
holds. From \cref{prop:LSE_optimality}, we know that $\bar{n}$ is finite. Consider $a \in R^c_{\epsilon}(S_{\bar{n}})\setminus S_{\bar{n}}$. Then we have that there exists $a' \in S_{\bar{n}}$ such that
$0 \leq g_i(a',x_0) - \epsilon - L_{\mathrm{a}}\normAny{a-a'}$ (see \cref{eq:reachability_operator}). Furthermore, $S_{\bar{n}-1}=S_{\bar{n}}$, means $a' \in  S_{\bar{n}-1}$. Hence, we also have $0 \leq u_{\bar{n}-1}(a,i)- L_{\mathrm{a}}\normAny{a-a'}$, which implies that, $a' \in \mathcal{G}_{\bar{n}-1}$ (\cref{sec:additional_defs} \cref{def:expanders}) and therefore, $w_{\bar{n}-1}(a',i) < \epsilon$. This implies that $0 \leq l_{\bar{n}-1}(a',i)- L_{\mathrm{a}}\normAny{a-a'}$. Therefore, according to \cref{eq:safeset}, $a \in S_{\bar{n}}$, which is a contradiction. Hence, $\Bar{R}^{\epsilon}_{c} (A) \setminus S_{\bar{n}} = \emptyset$. We can proceed similarly to show that $\Bar{R}^{\epsilon}_{c} (S_n) \setminus S_{\bar{n}} = \emptyset$. Since we have $\Bar{R}^{\epsilon}_{c} (A) \setminus S_{\bar{n}} = \emptyset$ and $\Bar{R}^{\epsilon}_{c} (S_n) \setminus S_{\bar{n}} = \emptyset$, we can conclude that $\Bar{R}^{\epsilon}_{c} (A) \cup \Bar{R}^{\epsilon}_{c} (S_n) \subseteq S_{\bar{n}}$.
\end{proof}
In \cref{lemma:explore_safe_region_after_ge} we have shown that for every set $A$ that we add to our safe set, we will explore its fully connected safe region in finite time. This is crucial because it allows us to guarantee that when we discover a new region during \gls{GE}, we explore it till convergence. Finally, we can now prove~\cref{thm:optimality}.
\optimality*
\begin{proof}
Since, $a^*$ is discoverable at iteration $\tilde{n}$, there exists a set $A^* \subseteq S_{\tilde{n}}$ such that $a^* \in \bar{R}^c_{\epsilon}(A^*)$. Furthermore, we have $\bar{R}^c_{\epsilon}(A^*) \subseteq \bar{R}^c_{\epsilon}(S_{\bar{n}})$ (\cref{lemma:explore_safe_region_after_ge}), therefore,  $a^* \in \bar{R}^c_{\epsilon}(S_{\bar{n}})$.
\cref{prop:LSE_optimality} shows that we can find the optimum in the safe region with $\epsilon$ precision in finite time $n^* \geq \tilde{n}$. 
Hence, there exists a finite integer $n^*$ such that
\begin{equation}
        f(\hat{a}_{n}) \geq f(a^*) - \epsilon, \quad \forall n \geq n^*.
\end{equation}
    with $\hat{a}_{n} =\argmax_{a \in S_{n}} l_{n}(a,0)$.

\end{proof}
\subsubsection{Requirements for Discovering Safe Sets with \gls{GE}} \label{discoverable_proofs}
In the previous section, we showed that if a safe global optimum $a^*$ is discoverable at some iteration $\tilde{n}$, we can then find it with $\epsilon$-precision.
In this section, we show that 
if for a parameter $a_{\gls{GE}}$ in $\mathcal{A}\setminus S_n$, we have backup policies for all the states in its trajectory, then  $a_{\gls{GE}}$ will be eventually added to our safe set of parameters. Finally, we conclude this section by showing that for many practical cases, $a^*$ fulfills the discoverability condition.

Now, we derive conditions that allow us to explore new regions/parameters during \gls{GE}.  To this end, 
we start by  defining a set of safe states $\mathcal{X}^{\text{s}}_{n}$, i.e., the states for which our boundary condition does not trigger a backup policy. 
\begin{definition}
The set of safe states 
$\mathcal{X}^{\text{s}}_{n}$ is defined as
\begin{equation}
    \mathcal{X}^{\text{s}}_{n}\coloneqq \bigcup_{(a',x') \in \mathcal{B}_{n}} \left\{ x \in \mathcal{X} \,\Bigl\lvert\, \normAny{x'-x} \leq \frac{1}{L_\mathrm{x}}\min_{i \in \mathcal{I}_g} l_n(a',i) -  \Xi, \right\}.
\end{equation}
\end{definition}
Intuitively, if a trajectory induced by a parameter being evaluated during \gls{GE} lies in $\mathcal{X}^{\text{s}}_{n}$, then the boundary condition will not be triggered for this parameter. 
Now we will prove that this set of safe states $\mathcal{X}^{\text{s}}_{n}$ is non-decreasing. This is an important property because it tells us that \gosafeopt continues to learn backup policies for more and more states.
\begin{lemma}
Let the assumptions from~\cref{thm:safety} hold. For any $n \geq 0$, the following property is satisfied for $\mathcal{X}^{\text{s}}_{n}$.
\begin{equation}
    \mathcal{X}^{\text{s}}_{n} \subseteq \mathcal{X}^{\text{s}}_{n+1}. \label{nondecreasing_stateset}
\end{equation}
\label{nondecreasing_property}
\end{lemma}
\begin{proof}
The lower bounds $l_n(a,i)$ are non-decreasing for all $i \in \mathcal{I}$ by definition (see \cref{alg:LSE} line~\ref{lst:line:update_S_lse} or \cref{alg:GE} line~\ref{lst:line:update_S_ge}).
Additionally, because we continue to add new rollouts to our set of backups, we have $\mathcal{B}_n \subseteq \mathcal{B}_{n+1}$ (see \cref{alg:LSE} line~\ref{lst:line:update_GP_lse} or \cref{alg:GE} line~\ref{lst:line:update_GP_ge}). For each $x \in \mathcal{X}^{\text{s}}_{n}$, there exists $(a_s,x_s) \in \mathcal{B}_n$, such that $l_n(a_s,i) - L_{\mathrm{x}} \left(\normAny{x-x_s} + \Xi\right) \geq 0$ for all $i \in \mathcal{I}_g$. Because $\mathcal{B}_n \subseteq \mathcal{B}_{n+1}$ and $l_{n+1}(a_s,i) \geq l_n(a_s,i)$,  $x \in \mathcal{X}^{\text{s}}_{n+1}$.
\end{proof}
Next, we state conditions under which a parameter $a_{\gls{GE}} \in \mathcal{A}\setminus S_n$ will be discovered during \gls{GE}, i.e., no backup policy would be triggered during \gls{GE}, in finite time.
\begin{lemma}\label{lemm:find_in_GE}
Consider any $n\geq 0$. Let $S_n$ be the safe set of parameters explored after $n$ iterations of \gosafeopt and $a_{\gls{GE}}$ a parameter in $\mathcal{A} \setminus S_n$. Further, let the assumptions from \cref{thm:optimality} hold and $\beta_n$ be defined as in \cite{DBLP:journals/corr/BerkenkampKS16}. If, for all $k\geq 0$, $x^{a_{\textrm{\gls{GE}}}}(k) \in \mathcal{X}^{\text{s}}_{n}$, where, $x^{a_{\textrm{\gls{GE}}}}(k)$ represents the state at time step $k$ for the system starting at $x_0$ with policy $\pi^{a_{\gls{GE}}}(\cdot)$,
then there exists a finite integer $\tilde{n} > n$, such that $a_{\gls{GE}} \in S_{\tilde{n}}$.
\end{lemma}
\begin{proof}
 Assume that there exists no finite integer $\tilde{n} > n$ such that $a_{\gls{GE}} \in S_{\tilde{n}}$. This would imply that  $a_{\gls{GE}} \in \mathcal{A} \setminus S_{\tilde{n}}$ for all $\tilde{n} > n$. Thus, because $a_{\gls{GE}}$ will never be a part of our safe set, it will never be evaluated during \gls{LSE}. However, from \cref{prop:LSE_optimality} we know that \gls{LSE} will converge in a finite number of iterations after which we will perform \gls{GE}. Since $a_{\gls{GE}}$ is not a part of the safe set, it can only be evaluated during \gls{GE}, where parameters outside of the safe regions are queried. 
 The parameter space $\mathcal{A}$ is finite and any parameter that was evaluated unsuccessfully, i.e., boundary condition was triggered, will be added to $\mathcal{E}$ and therefore not evaluated again (see \cref{S3_acquisition} and \cref{alg:GE} line~\ref{lst:line:fail_set_update}). 
 This implies that $a_{\gls{GE}}$ will be evaluated for some $n'$  with $n< n' < \tilde{n}$ (since $S_{n'} \subseteq S_{\tilde{n}}$, see \cref{nondecreasing_propertyS}).
 Furthermore, $\mathcal{X}^{\text{s}}_{n} \subseteq \mathcal{X}^{\text{s}}_{n'}$ (\cref{nondecreasing_property}) and, therefore, for all $k\geq 0$, $x^{a_{\textrm{\gls{GE}}}}(k) \in \mathcal{X}^{\text{s}}_{n'}$. 
 If when $a_{\gls{GE}}$ is evaluated, the experiment is unsuccessful, i.e., we were to trigger a backup policy, 
 this would imply that for some $k'$ and $i\in\mathcal{I}_g$, there is no $(a_s, x_s)\in\mathcal{B}_{n'}$ such that
 \begin{equation*}
     l_{n'}(a,i) \ge L_\mathrm{x}(\normAny{x^{a_{\textrm{\gls{GE}}}}(k')-x_s} + \Xi).
 \end{equation*}
 Thus, we had $x^{a_{\textrm{\gls{GE}}}}(k')\notin\mathcal{X}^{\text{s}}_{n'}$, which contradicts our assumption. Therefore, $a_{\gls{GE}} \in S_{n'+1} \subseteq S_{\tilde{n}}$ (\cref{nondecreasing_propertyS}), which is a contradiction. 
 Consequently, we have $a_{\gls{GE}} \in S_{\tilde{n}}$ after a finite number of iterations $\tilde{n}$.
 \end{proof}

\cref{lemm:find_in_GE} guarantees that, if the trajectory of parameter $a_{\gls{GE}}$ lies in $\mathcal{X}^{\text{s}}_{n}$, then it will eventually be added to our safe set, either during \gls{LSE} or during \gls{GE}. We utilize this result to provide a condition for the safe global optimum $a^*$ to be \emph{discoverable} at iteration $\tilde{n}$ by \gosafeopt. 
\begin{lemma}
Consider any $n\geq 0$. Let Assumptions~\ref{ass:safe_seed} --  \ref{ass:one_step_jump} hold, $\beta_n$ be defined as in \cite{DBLP:journals/corr/BerkenkampKS16}, and let $a^* \in \mathcal{A}$ be the safe global optimum. If there exists $a_{\gls{GE}} \in \mathcal{A} \setminus S_n$ such that 
\begin{enumerate}[label=(\roman*)]
    \item for all $k\geq 0$ and some $n\geq 0$, $x^{a_{\textrm{\gls{GE}}}}(k) \in \mathcal{X}^{\text{s}}_{n}$, where  $x^{a_{\textrm{\gls{GE}}}}(k)$ is the state visited by the system starting at $x_0$ under the policy $\pi^{a_{\gls{GE}}}(\cdot)$ at time step $k$, and
    \item $a^* \in \Bar{R}^{c}_{\epsilon}(\{a_{\gls{GE}}\})$,
\end{enumerate}
then $a^*$ is discoverable at iteration $\tilde{n} \geq n$ by \gosafeopt with probability at least $1-\delta$.
\label{lemm:discoverable}
\end{lemma}
\begin{proof}
Since, for all $k\geq 0$,
$x^{a_{\textrm{\gls{GE}}}}(k)$ $\in$ $\mathcal{X}^{\text{s}}_{n}$, there exists a finite integer $\tilde{n}>n$ such that $a_{\textrm{\gls{GE}}} \in S_{\tilde{n}}$  (Lemma~\ref{lemm:find_in_GE}). 
Furthermore, because $a^* \in \Bar{R}^{c}_{\epsilon}(\{a_{\gls{GE}}\})$, we can conclude that $a^*$ is discoverable at iteration $\tilde{n}$ (see \cref{def:discoverable_set}).
\end{proof}
The condition here is interesting because empirically, for many practical cases, it is fulfilled. Crucially, optimal or near-optimal parameters tend to visit similar states as other safe policies. We add rollouts from safe policies to our set of backups $\mathcal{B}_n$ and therefore have backup policies for their trajectories and other trajectories that lie close to them. Therefore, the trajectories of (near-)optimal parameters lie in $\mathcal{X}^{\text{s}}_{n}$ and in this case, the safe global optimum fulfills the discoverability condition from \cref{thm:optimality}. 
\section{Additional Information on Experiments} \label{exp_info}
For all our experiments in \cref{results}, we consider a controller in the operational space. The operational space dynamics of the end-effector are given by~\cite{robotics_handbook}
\begin{equation}
    u(x(t)) = \Lambda(q) \Ddot{s} + \Gamma(q,\Dot{q}) \Dot{s} + \eta(q),
    \label{OS_dynamics}
\end{equation}
where $s$ represents the end-effector position, $q$ the joint angles, and $\Lambda(q)$, $\Gamma(q,\Dot{q})$, $\eta(q)$ are nonlinearities representing the mass, Coriolis, and gravity terms, respectively. The state we consider is $x(t) = [s^T(t), \Dot{s}^T(t)]^T$. 
We apply an impedance controller:
\begin{equation}
u\left(x(t)\right) = -K (x-x_{\text{des}}(k)) 
    + \Gamma(q,\Dot{q}) \Dot{s} + \eta(q), 
    \label{controller}
\end{equation}
with $K$ being the feedback gain. 
The torque $\tau$ applied to each of the joints can be calculated via $\tau =  J^T u(x(t))$, with $J$ the Jacobian.

For our experiments, we can directly measure $g(a,x(k))$, where $k$ denotes a discrete time step. Therefore, instead of using $l_n(a_s,i)$ for the boundary condition in \cref{sec:UBC}, we take a lower bound over all the tuples in our set of backups, $\mathcal{B}_n$, i.e., $l_n(a_s,x_s,i)$, which could potentially reduce the conservatism of the boundary condition. Therefore, we 
define a \gls{GP} over the parameter and state space, which contains all the points from $\mathcal{B}_n$. The set $\mathcal{B}_n$ consists of rollouts from individual experiments, we typically add $50-100$ data points from each experiment to $\mathcal{B}_n$. As the data points of our \gls{GP} increase, inference becomes prohibitively costly.
To this end, we use a subset selection scheme to select a small subset of points $(a,x)$ from $\mathcal{B}_n$ at random with a probability that is proportional to $\exp{ (-\min_{i \in \gls{conset}} l^2_n(a,x,i))}$. Crucially, we want to retain points that have a small lower bound such that we have low uncertainty around these points. We perform this subset selection once our \gls{GP} has acquired more than $n_{\max}$ data points. Then, we select a subset of $m<n_{\max}$ points.
Lastly, as described in \cref{choosing_hyperparams}, for the boundary condition from \cref{sec:UBC}, we define the distances $d_u$, $d_l$ using covariances  $\kappa_u$, $\kappa_l$, respectively. Particularly, we pick $d_u$ such that $k(d_u) \geq \kappa_u$ for the stationary isotropic kernel $k$ that we use to model our \gls{GP} (same for $d_l$).
This makes the choice of $d_u$ more intuitive since it directly relates to the covariance function of our \gls{GP}.
\subsection{Simulation}\label{FR_sim}
For the simulation task, we determine the impedance $K$ using an infinite horizon \gls{lqr} parameterized via 
\begin{equation*}
Q = \begin{bmatrix}
            Q_r & 0\\
            0 & \kappa_d Q_r
\end{bmatrix}, 
Q_r = 10^{q_c} I_{3}, 
\end{equation*}
\begin{equation*}
R = 10^{r-2} I_{3}, 
A =\begin{bmatrix}
            0 & I_{3}\\
            0 & 0
\end{bmatrix}, 
B=\begin{bmatrix}
            0\\
            I_{3} 
            \end{bmatrix}.
\end{equation*}
The matrices $A$, $B$ are obtained assuming that we use a feedback linearization controller~\cite{robotics_handbook}. However, because we instead use an impedance controller, there are nonlinearities and imprecisions in our model. The parameters $q_c,r,\kappa_d$ are tuning parameters we would like to optimize. 
We define the desired path $x_{\mathrm{des}}(k)$ as 
\begin{align*}
    x_{\text{des}}(k) &= [p_0^T + \dfrac{k}{T_{traj}} (p_{\mathrm{des}}-s_0)^T, 0_{1x3}]^T \tag{8D task} \\
    x_{\text{des}}(k) &= x_{\text{des}}(\rho(k)) \tag{11D task}
\end{align*}
Here, $\rho(\cdot)$ is used to parameterize a cubic spline from $x_{0}$ to $x_{\text{target}}$.
The constraint  is:
\begin{align}
    \Bar{g}\left(x(t)\right) &= \dfrac{\normgeneral{s(t)-s_{\mathrm{des}}}{2}-\normgeneral{s(0)-s_{\mathrm{des}}}{2}}{\normgeneral{s(0)-s_{\mathrm{des}}}{2}} - \alpha , \quad \alpha = 0.08 \tag{8D task} \\
    \Bar{g}\big(x(t)\big) &= \zeta-\normgeneral{x(t)-x_{\mathrm{d}}\big(\rho(t)\big) }{2} \tag{11D task}.
\end{align}
The stage rewards, i.e., rewards received at each time step~\cite{Sutton1998}, are
\begin{align}
\begin{split}
       \mathcal{R}\left(x(t)\right) &= -\normgeneral{s(t)-s_{\mathrm{des}}}{2}^2/\normgeneral{s(0)-s_{\mathrm{des}}}{2}^2 \\
    &\quad\hspace{0.15em} - \frac{1}{25}\normgeneral{\tanh{\Dot{s}(t)}}{2}^2- \frac{1}{25}\normgeneral{\tanh{u\left(x(t)\right)}}{2}^2
       \end{split}
    \tag{8D task}\\
    \mathcal{R}\big(x(t)\big) &= -\nu_{\rho} (\rho(t)-1)^2 - \nu_{x} \normgeneral{x(t)-x_{\mathrm{d}}\big(\rho(t)\big) }{2} - \nu_{u}  \normgeneral{ \dfrac{u}{u_{\max}}}{2} \tag{11D task}.
\end{align}

Additionally, to encourage fast behavior in the eight-dimensional task, we only sample parameters for which the eigenvalues of $A-BK$ are less than a fixed threshold: $\textrm{eig}(A-BK) \leq -10$. Although this constraint is independent of the state, it can be evaluated before each experiment and parameters can be rejected if the criterion is not fulfilled. The value for $\kappa_d$ is heuristically set to $0.1$ for the first experiment (8D task). For the eleven-dimensional task, $\kappa_d$ is also tuned. 
In the eight-dimensional task, we observe that the underlying functions $f,g_i$ exhibit non-smooth behavior. Therefore we use the Mat\'ern kernel~\cite{Rassmussen} with parameter $\nu=\sfrac{3}{2}$ for our \gls{GP}. 
For the remaining tasks, we use the \gls{se} kernel.
\subsection{Hardware} \label{Hw_results}
For the hardware task, we define the subsequent objective and constraint functions: 
\begin{align*}
    \mathcal{R}\left(x(t)\right) &= - \normgeneral{s(t)-s_{\mathrm{des}}(t)}{2}, \\
    \Bar{g}\left(s(t)\right) &=  \normgeneral{s(t)-s_w}{{P,\infty}}-\psi,
\end{align*}
where $s_w$ represents the center of the wall in \cref{fig:hardware_path} and $\normgeneral{s-s_w}{{P,\infty}} \leq d_{w}$ defines the rectangular shaped outline around the wall and $\psi > d_{w}$.
\section{Sensitivity to \gls{LSE} and \gls{GE} Steps}\label{sec:ablation_study}
We analyze the sensitivity of the practical version of \gosafeopt with respect to $n_{\textrm{\gls{LSE}}}$ and $n_{\textrm{\gls{GE}}}$ on a simple one-dimensional toy example. The toy example consists of a one-dimensional system that has the following dynamics, stage reward, and constraint:
\begin{align*}
    s(k+1) &= 1.01 \sqrt{|s(k)|} -0.2 \sqrt{|a y(k)|}+v(k), \quad \text{ with } y(k) = s(k)+w(k)\\
    \mathcal{R}(s) &= -s^2, \\
    \bar{g}(s) &= s^2 - 0.81,
\end{align*}
with $v(k),w(k) \sim \mathcal{N}(0,10^{-4})$ and $a \in [-6, 5]$ the control parameter we would like to optimize. We consider a regulation problem, i.e.\ we start at $x_0 = 0$ and we would like our system to remain close to $x_0$. We run \gosafeopt for twenty iterations over twenty seeds for different $n_{\textrm{\gls{LSE}}}$ and $n_{\textrm{\gls{GE}}}$ values. 
\begin{figure}
    \begin{subfigure}{0.49\textwidth}
    \centering
    \includegraphics[width=0.85\textwidth]{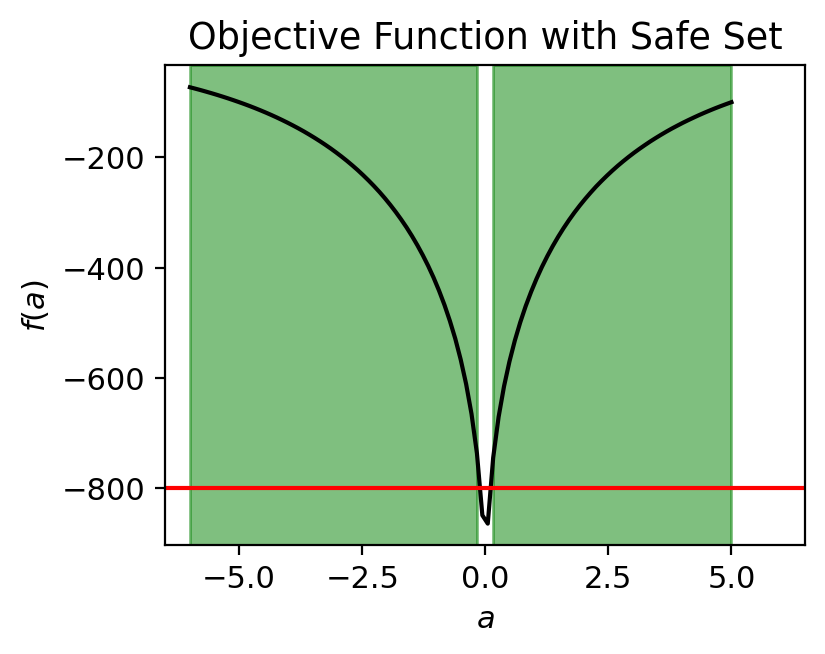}
    \caption{}
    \end{subfigure}
    \begin{subfigure}{0.49\textwidth}
    \centering
    \includegraphics[width=\textwidth]{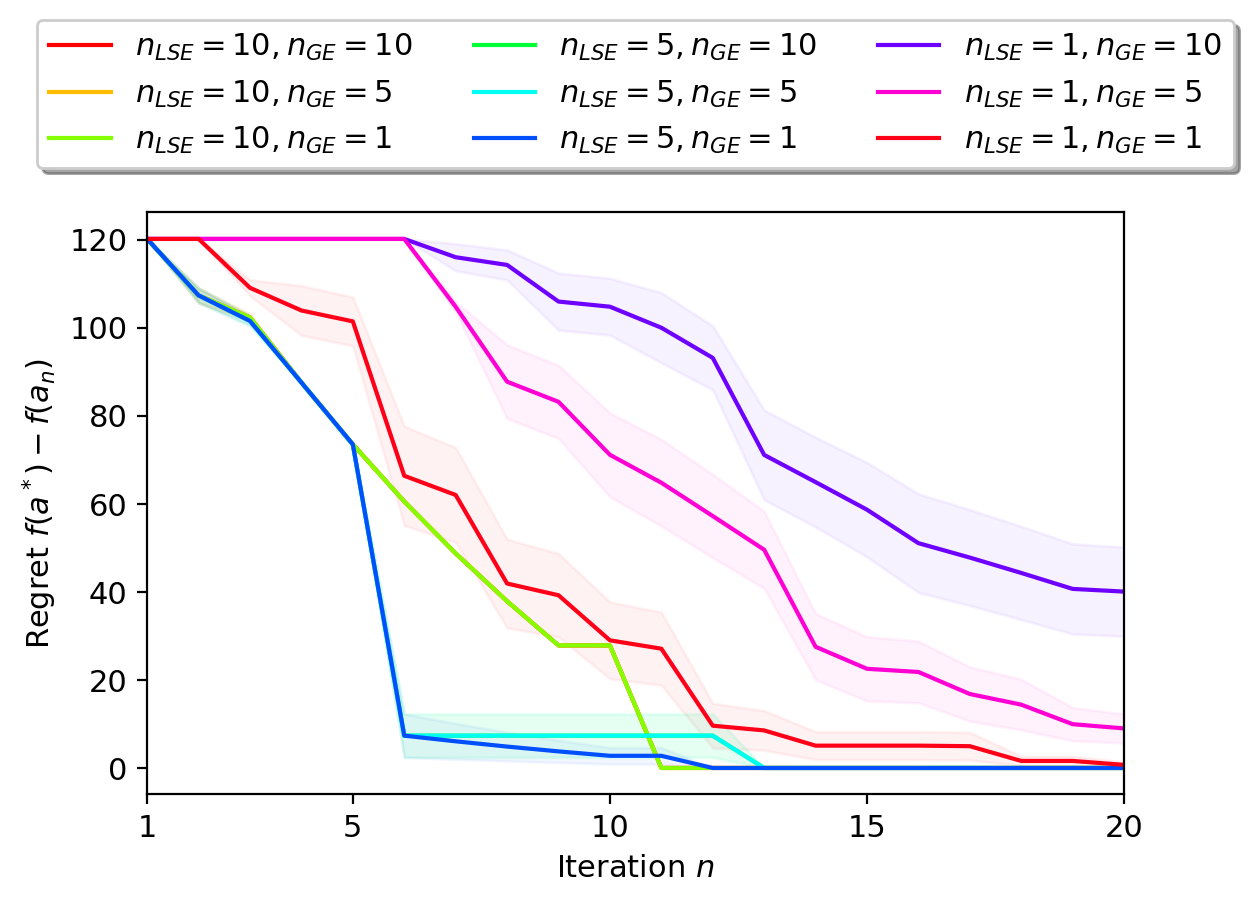}
    \caption{}
    \end{subfigure}
    \caption{Toy Example results. \capt{(a) Objective function with the safe set. (b) Performance of \gosafeopt for different $n_{\textrm{\gls{LSE}}}$ and $n_{\textrm{\gls{GE}}}$.}}
    \label{fig:result_ablation_study}
\end{figure}
\Cref{fig:result_ablation_study} depicts the safe set and the performance of \gosafeopt for different $n_{\textrm{\gls{LSE}}}$ and $n_{\textrm{\gls{GE}}}$. In this example, we have two disconnected safe sets and $a^*=-6$ is the global optimum. To show the advantage of global exploration, we initialize \gosafeopt in the safe region which does not contain $a^*$. For all our experiments, we obtain $100\%$ safety.  
Furthermore our results show, for $n_{\textrm{\gls{LSE}}} = 5$, we get faster convergence than for $n_{\textrm{\gls{LSE}}} = 10$. This is because we explore the region with the global optimum earlier. However, when we do not perform enough \gls{LSE} steps then global exploration also fails and we are stuck at a bad optimum (see for instance $n_{\textrm{\gls{LSE}}} = 1$ and $n_{\textrm{\gls{GE}}} = 10$). This simple example highlights the trade-off between local exploration  and global exploration steps. 
\section{Additional Definitions}
\label{sec:additional_defs}
In this section, we present some of the definitions from \safeopt for completeness.
\begin{definition} \label{def:expanders}
The expanders $\G$ are defined as $\G \coloneqq \{ a \in S_n\mid e_n(a) > 0\}$ 
    with $e_n(a) = |\{ a' \in \mathcal{A} \setminus S_n,  \exists i \in  \gls{conset}: u_n(a,i) - L_\mathrm{a} \normAny{a -a'} \geq 0 \} |$.
\end{definition}
\begin{definition}\label{def:maximizers}
The maximizers $\M$ are defined as $\M \coloneqq \{a \in S_n\mid u_n(a,0) \geq \max_{a'\in S_n} l_n(a',0) \}$.
\end{definition}
\section{Hyperparameters}\label{hyperparams}
The hyperparameters of our simulated and real-world experiments are provided in~\cref{tab:hyperparameter_sim}.
\begin{table}[htbp]
\caption{Table of Hyperparameters.}\label{hyperparameters_pos}
\begin{adjustbox}{max width=\linewidth}\begin{threeparttable}
\centering{
\begin{tabular}{c|cccccc}
\multicolumn{1}{l|}{} & \multicolumn{2}{c}{8D task Simulation} & \multicolumn{2}{c}{11D task Simulation} & \multicolumn{2}{c}{Hardware task}\\
              & \safeopt     & \gosafeopt     &  \safeoptswarm &\gosafeopt &\safeoptswarm &\gosafeopt                                                         \\ \midrule
Iterations    & $200$          & $200$    & $200$          & $200$     & $50$          & $50$                                                                  \\ 
$\beta_n^{\sfrac{1}{2}}$          & 4           & 4    &  3 & 3 & 3 & 3                                                              \\ 
$a$ lengthscale & $0.12$,$0.12$ & $0.12$,$0.12$   & $0.067$,$0.2$,$0.13$,$0.2$ & $0.067$,$0.2$,$0.13$,$0.2$   & $0.1$,$0.1$, $0.1$& $0.1$,$0.1$,   $0.1$                                                 \\ 
$\kappa$ for $f$ and $g$ & $1,1$ & $1,1$ & $1,1$ & $1,1$ & $1,1$ & $1,1$ \\ 
$\sigma$ for $f$ and $g$ & $0.1$,$0.1$ & $0.1$,$0.1$ & $0.1$,$0.1$ & $0.1$,$0.1$ & $0.05$,$0.3$ & $0.05$,$0.3$ \\ 
$x$ lengthscale & -           & \begin{tabular}[c]{@{}c@{}}$0.3$,$0.3$,$0.3$,\\ $2.5$,$2.5$,$2.5$ \end{tabular} 
& -           & \begin{tabular}[c]{@{}c@{}}$0.5$,$0.5$,$0.5$,\\ $0.6$,$0.6$,$0.6$, $10$ \end{tabular} & -           & \begin{tabular}[c]{@{}c@{}}$0.3$,$0.3$,$0.3$,\\ $0.5$,$0.5$,$0.2$ \end{tabular}
\\ 
$\epsilon$           & -           & $0.1$    &-   & $0.1$ &- & $0.01$                                                          \\ 
max \gls{LSE} steps  & -           & $30$       & -           & $100$ & -           & $20$                                                              \\ 
max \gls{GE} steps  & -           & $10$    & -           & $10$ & -           & $5$                                                              \\ 
$\kappa_l$      & -           & $0.90$      & -           & $0.90$  & -           & $0.90$                                                          \\ 
$\kappa_u$      & -           & $0.94$       & -           & $0.94$ & -           & $0.94$                                                         \\ 
$\eta_l$        & -           & $0.4$            & -           & $0.3$   & -           & $0.9$                                                       \\ 
$\eta_u$        & -           & $0.6$         & -           & $0.75$    & -           & $1.1$                                                     \\ 
$n_{\max}$      & -         & $1000$    & -         & $1000$  & -         & $1000$                                                              \\ 
$m$      & -         & $500$             & -         & $500$  & -         & $500$                                                     \\ \bottomrule
\end{tabular}}
\end{threeparttable}\end{adjustbox}
\label{tab:hyperparameter_sim}
\end{table}

\end{document}